\documentclass[11pt, a4paper, logo, copyright]{googledeepmind}

\usepackage[style=authoryear, backend=biber, natbib=true, maxcitenames=2, dashed=false]{biblatex}
\usepackage[capitalise]{cleveref}

\pdftrailerid{redacted}

\usepackage{kantlipsum, lipsum}
\usepackage{dsfont}
\usepackage{enumitem}

\newenvironment{compactenum}
  {\begin{enumerate}[noitemsep, topsep=0pt, parsep=0pt, partopsep=0pt]}
  {\end{enumerate}}
\usepackage{wrapfig}
\usepackage{gdm-colors}
\usepackage{subcaption}  
\usepackage[export]{adjustbox}  
\usepackage[algo2e, ruled,vlined]{algorithm2e}

\graphicspath{{figures/}}

\title{Building Production-Ready Probes For Gemini}

\correspondingauthor{conmy@google.com, janosk@google.com, neelnanda@google.com}


\reportnumber{001} 

\renewcommand{\today}{2026-01}

\author[$*$]{J\'{a}nos Kram\'{a}r}
\author[\ ]{Joshua Engels}
\author[\ ]{Zheng Wang}
\author[\ ]{Bilal Chughtai}
\author[\ ]{Rohin Shah}
\author[\ ]{Neel Nanda}
\author[$*$]{Arthur Conmy}

\affil[\ ]{Google DeepMind}
\affil[$*$]{Equal contributions to this work.}

\usepackage{mathtools}
\usepackage{dsfont}
\usepackage[dvipsnames]{xcolor}
\usepackage[colorinlistoftodos]{todonotes}
\usepackage{booktabs}
\usepackage{xfrac}
\usepackage{bbm}

\usepackage{algorithm}
\usepackage{algorithmicx}

\usepackage{xparse}
\usepackage{lipsum}
\usepackage{changepage}
\usepackage{enumitem}

\newcommand{\squishlist}{
   \begin{list}{$\bullet$}
    { \setlength{\itemsep}{0pt}      \setlength{\parsep}{3pt}
      \setlength{\topsep}{3pt}       \setlength{\partopsep}{0pt}
      \setlength{\leftmargin}{1.5em} \setlength{\labelwidth}{1em}
      \setlength{\labelsep}{0.5em} } }

\newcommand{\squishlisttwo}{
   \begin{list}{$\bullet$}
    { \setlength{\itemsep}{0pt}    \setlength{\parsep}{0pt}
      \setlength{\topsep}{0pt}     \setlength{\partopsep}{0pt}
      \setlength{\leftmargin}{2em} \setlength{\labelwidth}{1.5em}
      \setlength{\labelsep}{0.5em} } }

\newcommand{\squishend}{
    \end{list}  }

\newcommand{\myvec}[1]{\boldsymbol{#1}}

\newcommand{\two}{(2)}

\newcommand{\vq}{\myvec{q}}

\newcommand{\veevee}{\myvec{v}}
\newcommand{\vw}{\myvec{w}}

\newcommand{\vx}{\myvec{x}}

\newcommand{\vy}{\myvec{y}}

\newcommand{\vX}{\myvec{X}}

\DeclareMathAlphabet{\mathpzc}{OT1}{pzc}{m}{n}

\DeclareMathOperator*{\argmin}{arg\,min}
\usepackage{xcolor}

\usepackage{multirow}
\usepackage{float}  
\usepackage{longtable}  
\usepackage{xurl}

\begin{document}

\begin{abstract}
Frontier language model capabilities are improving rapidly. We thus need stronger mitigations against bad actors misusing increasingly powerful systems. Prior work has shown that activation probes may be a promising misuse mitigation technique, but we identify a key remaining challenge: probes fail to generalize under important production distribution shifts. In particular, we find that the shift from short-context to long-context inputs is difficult for existing probe architectures. We propose several new probe architectures that handle this long-context distribution shift.

We evaluate these probes in the cyber-offensive domain, testing their robustness against various production-relevant distribution shifts, including multi-turn conversations, long context prompts, and adaptive red teaming. Our results demonstrate that while our novel architectures address context length, a combination of architecture choice and training on diverse distributions is required for broad generalization. Additionally, we show that pairing probes with prompted classifiers achieves optimal accuracy at a low cost due to the computational efficiency of probes.

These findings have informed the successful deployment of misuse mitigation probes in user-facing instances of Gemini, Google's frontier language model. Finally, we find early positive results using AlphaEvolve \citep{novikov2025alphaevolve} to automate improvements in both probe architecture search and adaptive red teaming, showing that automating some AI safety research is already possible.
\end{abstract}

\maketitle

\section{Introduction}

In this paper, we describe our experience applying probes to detect cyber-offensive prompts given as input to Gemini 2.5 Flash \citep{flash}. We describe the challenges we faced and the solutions we arrived at as a case study for other frontier language model developers wishing to deploy probes as a misuse mitigation in production.

What do we mean by a misue mitigation? \textit{Misuse mitigations} are techniques that prevent malicious users from performing cyber-offensive, CBRN\footnote{Chemical, Biological, Radiological, and Nuclear} and other similar attacks using frontier large language models \citep{googledeepmind_fsf_v3_2025, anthropic_rsp_v2_2_2025, openai_preparedness_framework_v2_2025}. This risk is not theoretical: frontier models can already significantly increase the abilities of malicious users to perform these attacks \citep{Anthropic2025AIOrchestratedCyberEspionage}, and misuse mitigations have already been deployed by frontier AI companies to decrease this risk \citep{anthropic2025claude, deepthink}.\footnote{Anthropic: ``Our ASL-3 capability threshold for CBRN (Chemical, Biological, Radiological, and Nuclear)
weapons measures the ability to significantly help individuals or groups with basic
technical backgrounds (e.g. undergraduate STEM degrees) to create/obtain and deploy
CBRN weapons[...] we were unable to
rule out the need for ASL-3 safeguards.''. Google: ``Gemini 2.5 Deep Think continues the trend of increased model capabilities — it generates detailed technical knowledge of CBRN domains. It provides uplift in some stages of some harm journeys.''} One might hope that a sufficient mitigation would be to train LLMs to reject harmful queries, but unfortunately current training techniques are not robust enough \citep{nasr2025attacker}.


We thus focus on \textit{monitoring} in this paper: additional deployment time systems that aim to detect harmful user requests. In this work, we only study input monitoring techniques, although we note that training probes on model outputs is an important future direction in \Cref{secConclusion}.

In the main text we focus entirely on detecting the harmful and critical domain of cyber-misuse \citep{Anthropic2025AIOrchestratedCyberEspionage}, although we expect many findings to transfer between mitigation domains. Cyber-offensive capabilities are a particularly worthwhile domain for studying defenses because harmful prompts are very similar to common and valuable defensive cybersecurity requests, and also somewhat similar to even more common coding requests. Therefore, it is particularly difficult to avoid over-triggering while also preventing harm in the cyber domain. 

Recent work shows that large language model (LLM) classifiers can be an effective monitor \citep{sharma2025constitutional} for frontier models. But monitoring all interactions with another LLM can be extremely expensive, potentially doubling the cost if using a comparably capable model. A more cost-effective alternative is an activation probe: a small model trained on the internal hidden states of the model one plans to monitor. Activation probes are far cheaper than language model monitors because language model activations are already generated by the monitored model during a forward pass, so the only additional computational cost is running the probe itself. Indeed, \citet{cunningham2025cheap, cunningham2026} investigate probes and find evidence showing that they are a promising and cost-effective technique for misuse and jailbreak detection.

Overall, probes are indeed a cost-effective misuse mitigation, but there is a key problem: they are also fragile to distribution shifts. In particular, we find that probes trained on short-context data fail to generalize to important production distribution shifts, particularly long-context inputs. While training directly on long-context data is possible, we find it increases training costs by over an order of magnitude to achieve similar results due to memory bandwidth constraints (see \Cref{appInfra}), which in turn would significantly increase the costs of AlphaEvolve (\Cref{secAlphaEvolveResults}), for example. Distribution shifts like jailbreaks and multi-turn data are not fully addressed by our work, and hence challenges building robust and cheap defenses remain.


Our main technical contributions
are detailed investigation of four techniques that we investigate to address these distribution shift performance degradations. Below, we summarize these techniques and our main results from applying each one:
\vspace{-5mm}


\begin{enumerate}

\item \textbf{Improving probe architecture}: it is a difficult engineering challenge to train probes on long-context data, even with strong infrastructure (\Cref{appInfra}). Yet it is highly important that probes are performant on long-context data. We introduce a new probe architecture family, MultiMax, that has higher classifier accuracy on long context prompts compared to other probe architectures. In \Cref{secMainResults}, we show that MultiMax improves long context generalization compared to existing baselines such as mean-aggregated MLP probes (\Cref{figDistributionsComparisons}). 

\item \textbf{Automated probe architecture search}: In \Cref{secAlphaEvolveResults}, we also present results from an automated architecture search using AlphaEvolve \citep{novikov2025alphaevolve}, which discovers a probe that outperforms other baselines (bootstrap probability $> 0.95$, \Cref{secStatisticalSignificance}).

\item \textbf{Using a cascading classifier}: Large language models are inherently more robust to distribution shifts due to their generality, but are expensive to run. In \Cref{secCascadingClassifiers} we show that by only using the LLM when the probe is uncertain, we can create stronger classifiers than using either the probe or the LLM on its own, at a tiny fraction of the LLM's inference cost. For example, a probe-Flash cascade achieves lower test loss than Flash alone while using roughly 1/50th the inference cost (\Cref{figCascadingClassifiers}).

\item \textbf{The value of training probes across many different seeds}: We show that the high variance of different initialization of probes (particularly when training non-linear probes) can be exploited to find probes that fit both the validation and test set better (\Cref{subsecSeeds}), even though our train, validation and eval sets are disjoint (\Cref{secDatasetsAndMetrics}).


\end{enumerate}

These findings yield a set of recommendations for building high-quality probes. Using these recommendations, we find that our best probes achieve comparable performance to Gemini 2.5 Flash and Pro classifiers at a fraction of the cost. Probes also outperform 2.5 Flash Lite (\Cref{tabMainResults}). Furthermore, when we combine probes and LLMs into a cascading classifier, we can use the LLM less than 10\% of the time and still achieve a lower FNR than the LLM alone (\Cref{figCascadingClassifiers}). Our findings are not solely academic: they have directly informed GDM's recent successful deployment of cyber attack misuse mitigation probes into production. However, looking forward, our techniques do not significantly reduce the success rate of adaptive adversarial attacks; indeed, recent work argues that this is extremely challenging, if not impossible \citep{nasr2025attacker}. Our work can supplement other empirical studies \citep{cunningham2026} and user-level strategies \citep{shah2025approachtechnicalagisafety}.


We outline our notation and problem setup in \cref{secNotation}, the probing and classifier baselines we compare in \cref{secMethods}, the datasets we train and test on in \Cref{tabDatasets}, our evaluation metric in \Cref{subsecMetric}, and our main results in \cref{secResults}. We also study our new probe architectures and methodology on different models and domains in \Cref{appOtherDatasetsAndModels}.

\begin{figure}[htb]
\centering
\includegraphics[width=0.95\textwidth]{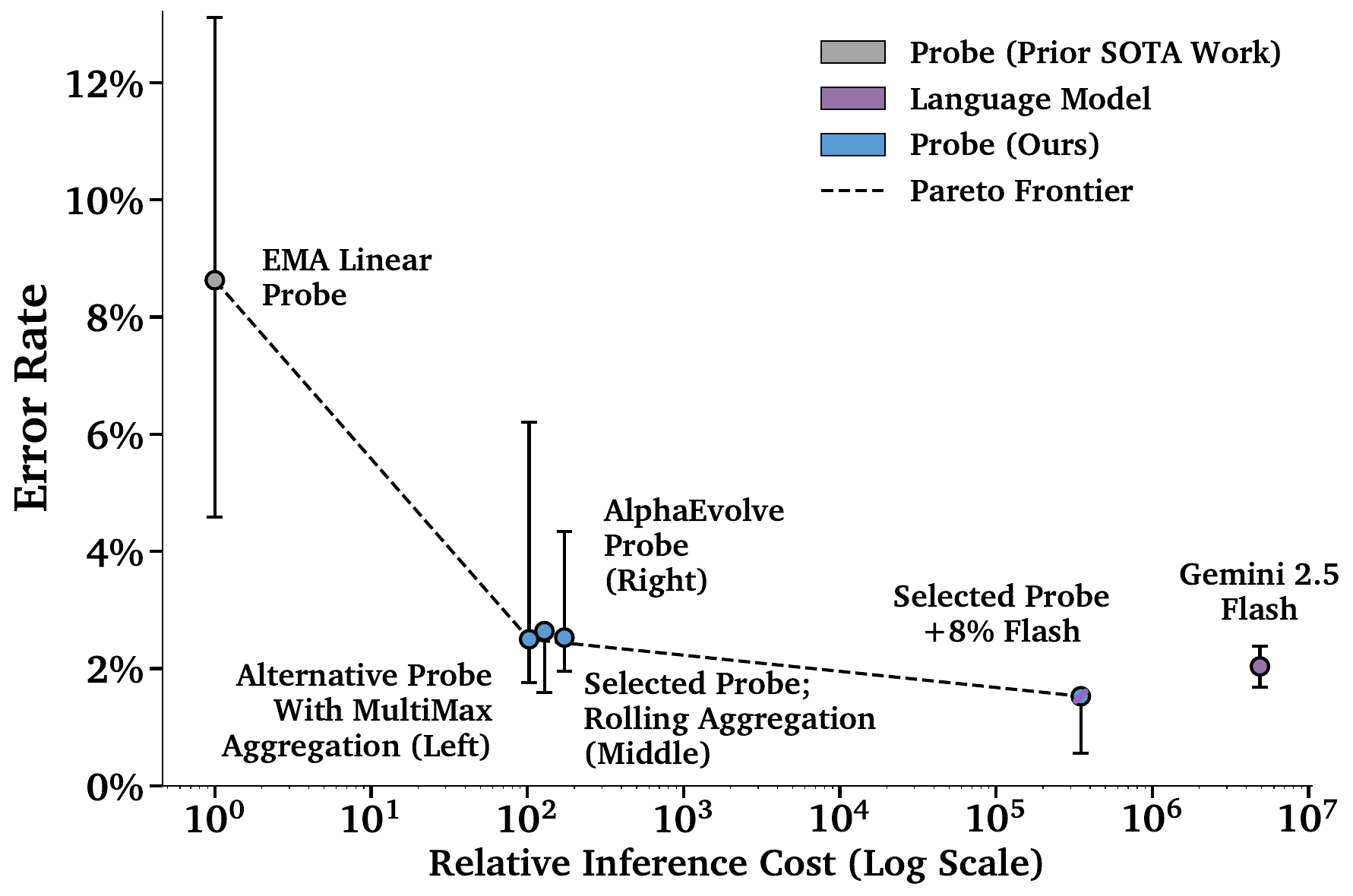}
\caption{We improve over prior activation probing work by improving probe architectures and training to achieve better performance than language models at over 10,000$\times$ lower cost. ``Selected Probe'' 
refers to our Max of Rolling Means Attention Probe (\Cref{secOurProbes}). Deferring to an LLM such as Gemini 2.5 Flash 8\% of the time further decreases error rate to a lower value than what can be achieved by solely using a probe or an LLM, by trusting the probe at extreme scores and delegating to the LLM in between (see \Cref{secOptimalFrontier}, though note wide error bars). \\
\textit{*As described in \Cref{subsecMetric}, for each method we select the decision threshold on a validation set by minimizing a weighted combination of FPR and FNR (i.e. error rate), with overtriggering FPR weighted most heavily. We report the same weighted error computed on test data. See \Cref{appErrorBars} for error bar methodology. The two points not on the Pareto frontier (Selected Probe and AlphaEvolve) have their costs artificially increased by 1.25$\times$ for visual separation; the costs of these three probes are in practice almost identical.}}
\label{figMainFigScatterCost}
\end{figure}


\section{Setup and Notation}
\label{secNotation}

We consider a deployed language model that receives prompts consisting of token sequences $S_i = (t_{i, 1}, \ldots, t_{i, n_i})$. Our goal is to classify sequences $S_i$ according to some criteria. We can ask a different LLM to perform this classification (see \Cref{secPromptedClassifier}). However, running an LLM for every user request is expensive, and we thus consider an alternative way of classifying $S_i$: training and running a small logistic regression classifier or neural network (i.e. a probe) trained on the deployed language model's hidden states (activations).

Formally, consider a training set of sequences $\{S_i\}$ with associated binary labels $\ell_i \in [0, 1]$. We pass each token sequence $S_i$ through the language model and harvest the dimension $d$ hidden states (one for each token) at a fixed layer $L$ (we use a layer from the middle of the model). Denote as $\vx_{i, j} \in \mathbb{R}^{d}$ the hidden activation at layer $L$ corresponding to $t_{i, j}$, and let $\vX_i \in \mathbb{R}^{d \times n_i}$ be the matrix of such hidden states corresponding to $t_i$. We train probes to minimize cross entropy loss on these sequences:

\begin{equation}
    f_p^* = \argmin_{f_p} \sum_{i} \text{BCE}(\sigma(f_p(\vX_i) + b), \ell_{i})
\end{equation}

where $b \in \mathbb{R}$ and the $\argmin$ is over all parameters of $f_p$ and $b$. We use \textit{full batch} gradient descent with AdamW, for 1000 epochs.\footnote{We found that it was easier to build infrastructure that loaded activations for training in once, rather than continually through training. See \Cref{appInfra}. We additionally only ever observed zero or mild overfitting on validation sets despite many epochs.}

\noindent During inference of the target model on a new sample $S^*_i$ (which could be from a different distribution to all $S_i$), we perform a single forward pass and run the trained probe on the activations $\vX^*_i$. If the probe score is above a \textit{threshold} $\tau$, we can run a misuse mitigation instead of naively continuing generation.
As a slight abuse of notation, we define $f_p(\cdot, \tau)$ to be the classification function from combining the probe $f_p$ with the threshold $\tau$:

\begin{equation}
    f_p(S_i, \tau) = \mathds{1}_{\{\sigma(f_p(\vX_i) + b) > \tau\}}
\end{equation}

where $\mathds{1}$ is the indicator function. This produces a classification decision, just as the prompted classifier does.

\section{Our Classifiers}
\label{secMethods}

\begin{figure}[ht!]
	\centering
	\includegraphics[width=\columnwidth]{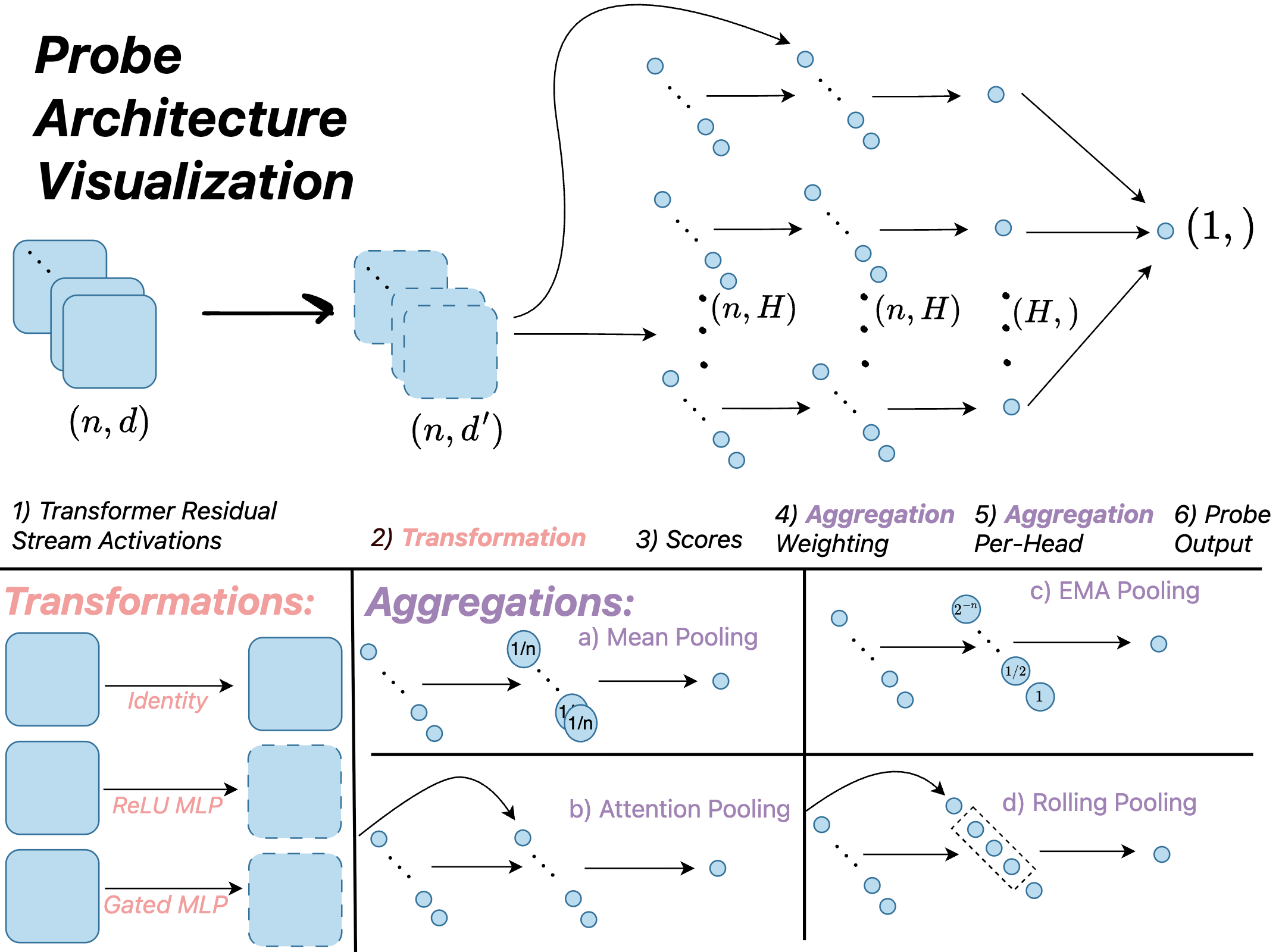}
	\caption{Different probing classifiers that we compare in \Cref{secMethods}. All our probing classifiers can be composed into six states 1) -- 6) as illustrated. Residual stream activations of undergo a 2) {\color[HTML]{F08080} Transformation} per-position. These are then processed via 4) -- 5) {\color[HTML]{9B59B6} Aggregation} which ends up producing a single scalar score at step 6).  Many existing probing classifiers such as linear probes, exponential moving average aggregation probes and attention probes fit into this framework.}
	\label{figClassifiers}
\end{figure}

We compare three categories of classifiers: baseline probe architectures from prior work, our novel probe architectures, and LLM-based classifiers. All methods are evaluated together in \Cref{secResults}; we separate them here for clarity. \Cref{figClassifiers} visualizes all architectures.

\subsection{Baseline Probe Architectures}
\label{secBaselineProbes}

We first describe probe architectures from prior work that serve as our baselines. 


\subsubsection{Linear Probes} Linear probes were first introduced by \cite{alain2016understanding}. Given $\vw \in \mathbb{R}^{d}$, mean pooled linear probes are defined as:

\begin{equation}
\label{eqn:mean_probe}
f_{\text{Linear}}(S_i) = \frac{1}{n_i} \sum_{j=1}^{n_i} \vw^T \vx_{i, j}
\end{equation}

so are very simple in terms of the steps 1) -- 6) in \Cref{figClassifiers}. Spelled out, for linear probes the transformation is just the identity map, $H=1$, the scores are a simple linear map, and aggregation is just the mean.


\subsubsection{Exponential Moving Average (EMA) Probes}  EMA probes were proposed by \cite{cunningham2025cheap} as a way to improve probe generalization to long contexts. To use an EMA probe, we first train a mean probe $f_{linear}$ as in \Cref{eqn:mean_probe}. Then, at inference time, we compute an exponential moving average at each index $j = 1, \ldots, n_i$:

\newcommand{\EMA}{\texttt{EMA}}

\begin{equation}
    \EMA_0 = 0 ; \qquad \EMA_{j} = \alpha f_{linear}(\vx_{i, j})+ (1 - \alpha)  \EMA_{j - 1} 
\end{equation}
for $\alpha \in (0, 1)$. Following \cite{cunningham2025cheap}, we use $\alpha = 0.5$. We then take the maximum over EMA scores to get $f_{\EMA}(S_i) = \max_{j} \EMA_j$ which is the final probe score.

\subsubsection{MLP Probes}
\label{subsubsecMLP}
MLP probes (e.g. \citet{breakers}) are the same as linear probes except that we apply an MLP to the pooled or single token activation vector before multiplying by $\vq$. Using the definition of an $M$ layer (ReLU) MLP as 
\begin{equation}
    \label{eqn:mlp_def}
    \text{MLP}_{M}(\vX) = A_1 \cdot \text{ReLU} (A_2 \cdot \text{ReLU} (\ldots \text{ReLU} (A_M \cdot \vX) \ldots ))
\end{equation}
an $M$-layer mean pooled MLP probe is formally defined as:

\begin{equation}
    f_{\text{MLP}}^M(S_i) = \frac{1}{n_i} \sum_{j=1}^{n_i} \text{MLP}_M(\vx_{i, j})
\end{equation}

with the dimensions of $A_1, \ldots, A_M$ chosen such that the final result is a scalar.

\subsubsection{Attention Probes}
\label{subsubsecAttention}
The single head attention probe was first introduced by \cite{kantamneni2025sparse}, further used by \cite{mckenzie2025detecting}, and expanded to use multiple heads by \cite{shabalin2025attention}. In our work, we first pass activations through a common MLP before computing attention. Specifically, we define:
\begin{equation}
    \vy_{i,j} = \text{MLP}_M(\vx_{i,j})
\end{equation}
\noindent where $\text{MLP}_M$ is an $M$-layer MLP as defined in \Cref{eqn:mlp_def}. An $H$-headed attention probe is then defined as:
\begin{equation}
\label{eqnAttentionProbe}
    f_{\text{Attn}}(S_i) = \sum_{h=1}^{H} \frac{\sum_{j=1}^{n_i} \exp(\vq_h^\top \vy_{i,j}) \cdot (\veevee_h^\top \vy_{i,j})}{\sum_{j=1}^{n_i} \exp(\vq_h^\top \vy_{i,j})}
\end{equation}
\noindent where $\vq_h, \veevee_h \in \mathbb{R}^{d'}$ are learned query and value vectors for head $h$, and $d'$ is the output dimension of the MLP. Note that this is the first probe where the aggregation weights are a function of the activations themselves rather than constants, illustrated by the looping upper arrow in \Cref{figClassifiers}.

One might naively think that attention probes require an expensive recomputation of the softmax function for every new token during generation. But this is false, and as an additional contribution we present an inexpensive inference algorithm for attention probes in \Cref{appAttentionProbeInference}.

\subsection{Our Probe Architectures}
\label{secOurProbes}

We now present our novel probe architectures designed to address the limitations of baseline methods, particularly for long-context generalization.

\subsubsection{MultiMax Probes}
\label{subsubsecMultimax}
We find that the attention probe's softmax weighting suffers from overtriggering on long contexts (see \Cref{secMainResults}). To address this, we propose the MultiMax architecture, which replaces the softmax with a hard max at inference time (though not always during training). Using the same MLP-transformed activations $\vy$ as above:
\begin{equation}
    f_{\text{MultiMax}
    }(S_i) = \sum_{h=1}^{H} \max_{j \in [n_i]} \left[ \veevee_h^\top \vy_{i,j} \right]
\end{equation}
\noindent where $\veevee_h \in \mathbb{R}^{d'}$ are learned value vectors for each head $h$. Unlike attention probes, MultiMax selects the single highest-scoring token per head rather than computing a weighted average, which prevents dilution of the signal when harmful content appears in a small portion of a long context.

\subsubsection{Max of Rolling Means Attention Probe}
\label{subsubsecRollingMeans}
In prior work, \citet{mckenzie2025detecting}\footnote{Max-pooling over local windows is a well-established technique in text classification, with precedent in convolutional approaches \citep{kim2014convolutionalneuralnetworkssentence, collobert2011naturallanguageprocessingalmost}.} introduce the Max of Rolling Means probe which takes the max over all mean scores of all possible windows of contiguous tokens of a fixed length in a sequence. In our work, we combine this with an Attention Probe, using the lessons learnt from the MultiMax probe work. Specifically, we use 10 attention heads by default and compute attention-weighted averages within sliding rectangular windows of fixed width $w$ (we use $w=10$ as default). For each window ending at position $t$, we compute:
\begin{equation}
    \bar{v}_t = \frac{\sum_{j=t-w+1}^{t} \alpha_j \cdot v_j}{\sum_{j=t-w+1}^{t} \alpha_j}
\end{equation}
\noindent where $\alpha_j = \text{softmax}(\vq^\top \vy_j)$ are the attention weights and $v_j = \veevee^\top \vy_j$ are the per-token values. The final output is $\max_t \bar{v}_t$. In this work, we explore using this aggregation at evaluation time, in addition to the simpler MultiMax aggregation described in \Cref{subsubsecMultimax}. 

\subsubsection{AlphaEvolve Architectures} We additionally experiment with running AlphaEvolve~\citep{novikov2025alphaevolve} to generate novel probing architectures. We start with an attention probe (\Cref{subsubsecAttention}) as our seed model\footnote{Since this is already an existing architecture (\Cref{subsubsecAttention}).} and use weighted FPR and FNR as our optimization target (we use somewhat of a different weighting to either \Cref{eqnTradeoff} and \Cref{appDifferent1250Weighting} due to using AlphaEvolve early in this research project, see \Cref{subappAlphaEvolveSetup}). We choose two of the architectures AlphaEvolve discovered as additional methods to compare to and include pseudocode for these two architectures in \Cref{subappAlphaEvolvePseudocode}.

\subsection{LLM-Based Classifiers}
\label{secLlmClassifiers}

Finally, we describe classifiers that use language models, which serve as both strong baselines and components of our cascading systems.

\subsubsection{Prompted Language Model}
\label{secPromptedClassifier}
A prompted language model $f_{LM}$ consists of a (prompt, language model) pair. To classify a sequence $S_i$, we append $S_i$ to the prompt and pass it to the language model, and then parse the language model's response. We use a fixed detection prompt provided by colleagues and consider Gemini 2.5 Flash Lite, Gemini 2.5 Flash, and Gemini 2.5 Pro as the language model. This method directly returns a classification decision, i.e. we do not need to pass it through a $C$ thresholding function. In \cref{appHillclimbingPrompt} we experiment with eight other prompts, but do not see a significant improvement, and so keep our original prompt.

\subsubsection{Cascading Classifier} Cascading classifiers were also studied by \cite{cunningham2025cheap} but have a long history in machine learning \citep{viola2001rapid}. We consider two level cascading classifiers, with level one set to be a probe $f_p$ and level two set to be a more expensive prompted language model $f_{LM}$. Then the cascading classifier is defined with two thresholds $\tau_1$ and $\tau_2$ as
\begin{equation}
    f_{cc}(S_i, \tau_1, \tau_2) = \begin{cases} 
    0 &\text{if $f_{p}(S_i) \le \tau_1$} \\
    f_{LM} &\text{if $\tau_1 < f_{p}(S_i) < \tau_2$}\\
    1 &\text{if $f_{p}(S_i) \ge \tau_2$}
    \end{cases}
\end{equation}
By varying the percent of samples spanned by $\tau_1$ and $\tau_2$ we can vary the additional computational cost incurred by the cascading classifier over just running the probe.

\section{Datasets and Metrics}
\label{secDatasetsAndMetrics}

\definecolor{lightgreen}{HTML}{C8E6C9}
\definecolor{lightred}{HTML}{FFCDD2}

\newcommand{\yes}{\cellcolor{lightgreen}yes}
\newcommand{\no}{\cellcolor{lightred}no}
\newcommand{\FPRscot}{\ensuremath{\texttt{FPR}_{\texttt{SC[OT]}}}}
\newcommand{\FPRschn}{\ensuremath{\texttt{FPR}_{\texttt{SC[HN]}}}}
\newcommand{\FPRmthn}{\ensuremath{\texttt{FPR}_{\texttt{MT[HN]}}}}
\newcommand{\FPRlcrt}{\ensuremath{\texttt{FPR}_{\texttt{LC[RT]}}}}
\newcommand{\FNRsca}{\ensuremath{\texttt{FNR}_{\texttt{SC[A]}}}}
\newcommand{\FNRmta}{\ensuremath{\texttt{FNR}_{\texttt{MT[A]}}}}
\newcommand{\FNRlcsa}{\ensuremath{\texttt{FNR}_{\texttt{LC[A]}}}}
\newcommand{\FNRscj}{\ensuremath{\texttt{FNR}_{\texttt{SC[J]}}}}
\newcommand{\FNRart}{\ensuremath{\texttt{FNR}_{\texttt{SC[ART]}}}}
\begin{table}[tb]
	\centering
	\caption{Errors we seek to minimize with our trained probe. For shorthand reference later on we label the error on each dataset as \textit{FPR} (for false positive rate) or \textit{FNR} (for false negative rate) with subscript of \texttt{prompt\_structure[prompt\_type]}}
	\begin{tabular}{llccc}
		\toprule
		\textbf{Description} & \textbf{Var} & \textbf{Train Split} & \textbf{Validation Split} & \textbf{Test Split} \\
		\midrule
		\multicolumn{5}{l}{\textit{Minimize false positive rate on...}} \\
		\hspace{2mm} short context overtriggering & \FPRscot & \yes & \yes & \yes \\
        \hspace{2mm} short context ``hard negatives'' & \FPRschn & \yes & \no & \yes \\
        \hspace{2mm} multi-turn (also ``hard negatives'') & \FPRmthn & \yes & \no & \yes \\
		\hspace{2mm} long context random traffic & \FPRlcrt & \no & \no & \yes \\
        \midrule
		\multicolumn{5}{l}{\textit{Minimize false negative rate on...}} \\
		\hspace{2mm} short context attacks & \FNRsca & \yes & \yes & \yes \\
        \hspace{2mm} multi-turn attacks & \FNRmta & \yes & \no & \yes \\
		\hspace{2mm} long context attacks & \FNRlcsa & \no & \no & \yes \\
        \hspace{2mm} pre-existing jailbreaks & \FNRscj & \no & \no & \yes \\
        \hspace{2mm} automated red teaming & \FNRart & \no & \no & \yes \\
		\bottomrule
	\end{tabular}
    \label{tabDatasets}
\end{table}

We focus on a single misuse domain in this paper: cyber misuse. Our breakdown of which datasets we train on, which we validate on, and which we test on is outlined in \Cref{tabDatasets}. We briefly describe each dataset in \Cref{tabDatasetDescriptions}, sorted into datasets with benign examples on which we want our probe to classify with logits less than 0 (and thus on which we want to minimize the false positive rate), and datasets with true cyber attacks on which we want our probe to classify with logits greater than 0 (and thus on which we want to minimize false negative rate). Importantly, no prompts are shared between any two of the train, validation, and test splits. Furthermore, all test sets represent distribution shifts from the training and validation distributions.

We intentionally train probes on a subset of distribution shifts, and in particular not on long context, since training on long-context dramatically increases infrastructure burden and compute. This informs our decision of which training sets are included in \Cref{tabDatasets}. Additionally, we use a different distribution for validation and test sets to ensure that thresholds are not optimized on the same distribution as the test set.

Note: we use the term ``overtriggering'' to refer to benign (non-cyber) datasets on which our classifiers should not fire, including PII-redacted user traffic, capability evaluation data and synthetic prompts curated for unrelated purposes.

\begin{small}
\begin{longtable}{@{}p{0.18\textwidth} p{0.78\textwidth}@{}}
\caption{Detailed descriptions of each dataset. See \Cref{appDatasetStats} for metadata including dataset sizes.}
\label{tabDatasetDescriptions} \\
\toprule
\textbf{Dataset} & \textbf{Description} \\
\midrule
\endfirsthead
\multicolumn{2}{c}{\tablename\ \thetable{} -- \textit{Continued from previous page}} \\
\toprule
\textbf{Dataset} & \textbf{Description} \\
\midrule
\endhead
\midrule
\multicolumn{2}{r}{\textit{Continued on next page}} \\
\endfoot
\bottomrule
\endlastfoot
\multicolumn{2}{l}{\textit{Benign (minimize FPR)}} \\
\cmidrule{1-2}
Short Context\newline Overtriggering & Short benign language model prompts. This dataset combines random coding traffic samples, random chat traffic samples with PII removed, and a mixture of Gemini post-training datasets. Specifically, we train on random traffic from one Google surface where Gemini is served, and use traffic from a different Google surface in the validation and evaluation sets (both surfaces have PII scrubbed). In the validation set, we use exclusively non-coding-related overtriggering prompts, and in the test set we exclusively use coding prompts, ensuring a distribution shift. \\
\cmidrule{1-2}
Short Context\newline ``Hard Negatives'' & Short benign cybersecurity requests that superficially resemble attack queries. These include requests to understand historical security incidents (e.g., ``Explain how the 2017 Equifax breach exploited the Apache Struts vulnerability''), requests to review code for vulnerabilities as part of secure development practices, and queries from learners using educational security platforms. This dataset is curated and labeled by cybersecurity domain experts. \\
\cmidrule{1-2}
Multi Turn\newline ``Hard Negatives'' & Benign, synthetically-generated, multi-turn conversations that discuss cybersecurity topics. These conversations are designed to mimic multi-turn attacks where malicious intent only becomes apparent over several turns; however, in these hard negative examples, the conversation never becomes malicious. This helps train classifiers to avoid firing on legitimate, evolving conversations about security. We use synthetically created examples for training, bootstrapping single turn conversations together. \\
\cmidrule{1-2}
Synthetic Long Context\newline Random Traffic & Long context prompts generated from coding user traffic with PII removed (on average >900k tokens, though at most 1,040,000 tokens due to Gemini context length limits). These random traffic coding queries come from the \textit{Short Context Overtriggering} dataset evaluation set, so there is no leakage between train and test data. More specifically, we take the coding related queries, get Gemini to recall several GitHub repositories with related content to the query, filter for those with permissive licenses, and create a near maximal length prompt with these codebases as either prefix or the suffix of the request. \\
\midrule
\multicolumn{2}{l}{\textit{Attack (minimize FNR)}} \\
\cmidrule{1-2}
Short Context Attacks & Short (< 10k tokens) instances of cyber attack prompts, e.g. ``How do I execute a SQL injection attack''. Curated and labeled by cyber-security domain experts. \\
\cmidrule{1-2}
Long Context\newline Attacks & Long context prompts generated from harmful short-context Cyber queries (which are a different set to those used in all other datasets here). We use a similar long context generation pipeline to the ``Long Context Random Traffic'' dataset, except insert harmful requests as a prefix to the code (i.e. at the start of the prompt), suffix to the code (i.e. at end of the prompt), as well as just before the final repository (so the request is interleaved with the context). \\
\cmidrule{1-2}
Multi Turn\newline Attacks & Synthetically-generated, multi-turn conversations that begin with benign or ambiguous cybersecurity-related prompts but escalate over several turns to become explicitly malicious. This dataset simulates attackers attempting to bypass single-turn defenses by gradually building a harmful context, training the classifier to detect threats that only become apparent over an entire conversation. Generated using the same experts as the \textit{Multi Turn ``Hard Negatives''}. \\
\cmidrule{1-2}
Pre-existing\newline Jailbreaks & We take a list of jailbreaks, i.e. functions that map strings to strings such as appending a suffix as in the GCG paper \citep{zou2023universal}, and apply them to all of the harmful cyber prompts in the \textit{Short Context Attacks} evaluation set. \\
\cmidrule{1-2}
Adaptive Red\newline Teaming (ART) & This jailbreak dataset is constructed as follows: for each attack in the short context attacks dataset, we ask Gemini 2.5 Pro to construct a narrative prompt that encodes the attack as a natural part of the prompt. Importantly, this process could in principle transform a malicious prompt into a benign one; we guard against this by using an autorater to verify that the model's response is actually harmful (see \Cref{appArtSuccessDefinition} for details). Our initial version of the ART dataset was much more complicated. It used an adaptive jailbreak finding method modeled off of Prompt Automatic Iterative Refinement (PAIR; \citet{chao2025jailbreaking}) to find semantic jailbreaks by iteratively prompting an attacker model to generate a jailbreak for a given query and giving it feedback on whether it succeeded. However, we found that this complicated method was not significantly better than the single shot baseline described above; see \cref{secArtAnalysis}. \\
\end{longtable}
\end{small}

\subsection{Metric}
\label{subsecMetric}

There are many different datasets in \Cref{tabDatasets}, and for all methods besides the prompted classifier methods we must choose a threshold $\tau$ that trades off against all of these metrics for use on the test splits. We make this choice by choosing a $\tau^*$ that minimizes the following weighted error metric on all validation datasets from \Cref{tabDatasets}:

\begin{equation}
\label{eqnTradeoff}
\tau^* = \argmin_\tau \Bigg[ 
    \underbrace{
        \frac{\sum {w_i} \cdot \texttt{ERROR}_i}{\sum w_i} 
    }_{\text{Validation Loss}} 
\Bigg] 
\qquad 
\texttt{w}_i = \begin{cases} 
    5  & \text{if $\texttt{ERROR}_i$ is \texttt{FNR}}\\ 
    2  & \text{if $\texttt{ERROR}_i$ is hard negative \texttt{FPR}}\\ 
    50 & \text{if $\texttt{ERROR}_i$ is overtriggering \texttt{FPR}}
\end{cases}
\end{equation}

This equation allows us to weight performance on datasets differently. Specifically, we weight overtriggering false positives 10 times higher than false negatives and 25 times higher than hard negative false positives. These multipliers reflect that true negatives are very rare and that we want to almost never classify benign overtriggering prompts as malicious, whereas occasionally classifying hard negatives as malicious is acceptable. See \Cref{appDifferent1250Weighting} for more discussion and results with alternative weightings.

We also use \Cref{eqnTradeoff} to calculate a weighted test loss on all \textit{test} datasets from \Cref{tabDatasets}, which we use as the main metric when we need to compare methods along a single axis (e.g. \Cref{figCascadingClassifiers}). In other words, to compute a test loss, we take the weighted average of all FPRs and FNRs on the test set instead of validation set.

\subsection{Initialization Seeds}
\label{subsecSeeds}

We train all probes in \Cref{tabMainResults} on 100 different seeds, measure their validation loss on the validation sets in \Cref{tabDatasets}, and then choose the seed that has the lowest validation loss. We then report the test loss for this chosen seed.

\begin{figure}[tb]
    \centering
    \includegraphics[width=\columnwidth]{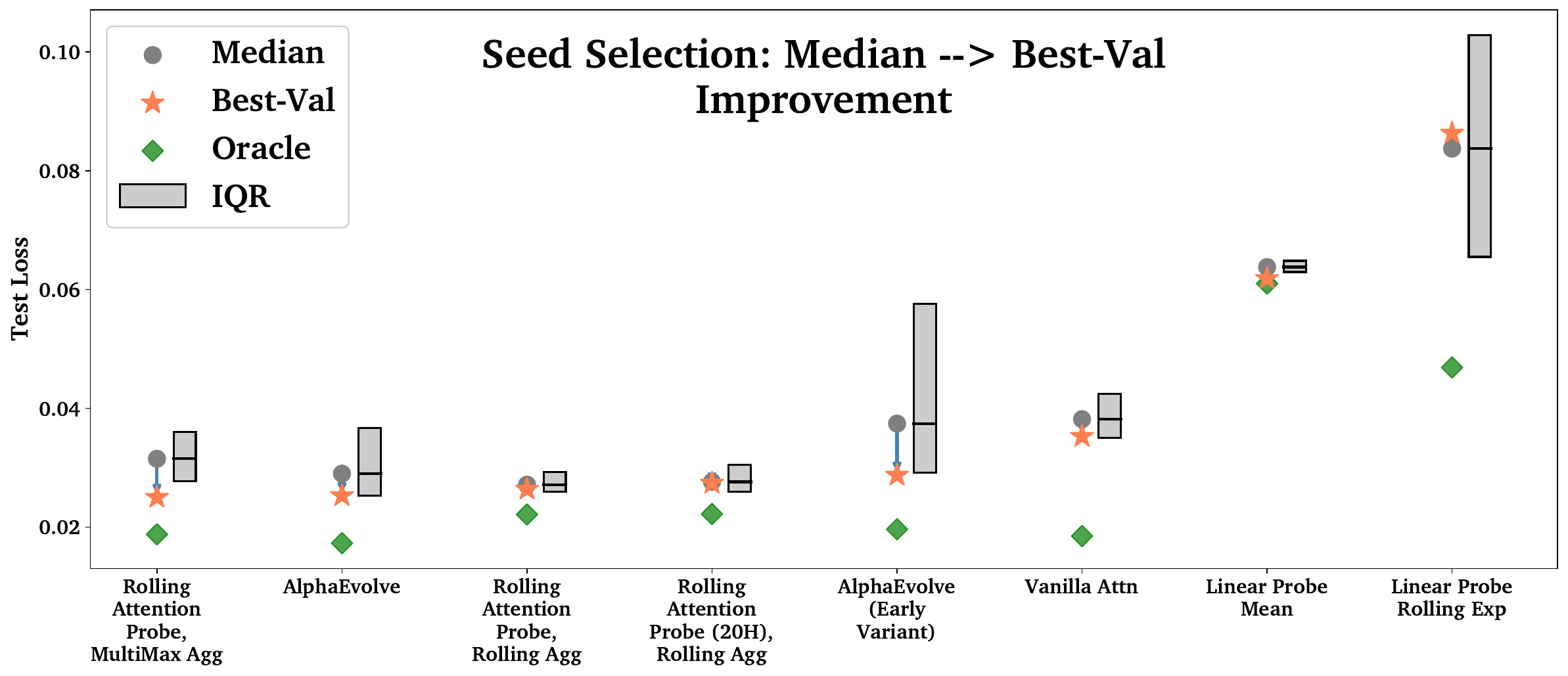}
    \caption{Effect of seed selection on test loss. Arrows show improvement from median (gray) to best-validation-selected (coral star) test loss. Green diamonds show the oracle (best possible seed). Some MultiMax aggregations have very large IQR and are shown in \Cref{figAppendixSeedSelection} to not distort this figure.}
    \label{figSeedSelection}
\end{figure}

In \Cref{figSeedSelection} we show the effect of this seed selection procedure on test loss across architectures. To quantify the impact of choosing the best seed on validation, we compare the test loss of the best-validation-loss seed to the test loss of the median-validation-loss seed for each architecture. Across architectures, seed selection via lowest validation loss reduces test loss by 0.008 on average, though the median reduction is only 0.003. In contrast, architecture choice provides a much larger gain. Comparing the median-validation-loss seeds, the best architecture achieves 0.025 test loss while the Linear Probe Mean achieves 0.064, over 13$\times$ larger than the median seed selection gain. This result suggests that practitioners should prioritize architecture search over extensive seed tuning. However, researchers should still sweep over multiple seeds to reduce variance, as individual seed performance can vary substantially and does still provide a small uplift. This is a finding consistent with prior work on seed sensitivity in classifier finetuning \citep{risch-krestel-2020-bagging} and neural network training more broadly \citep{picard2023torch}.

We also note that the best possible seed on the test set is frequently much better than the seed we choose via our validation dataset. It is possible that with a better validation data mix, the best-val seed would move closer to the best possible seed.

\section{Results}
\label{secResults}

\subsection{Main Cyber Probing Results}
\label{secMainResults}

In \Cref{tabMainResults} we show our evaluation of a series of probes and prompted classifiers across all datasets in increasing order of test set error. Our key takeaways are that:

\begin{compactenum}
\item Existing probe architectures can be significantly improved upon.
\item No single method dominates across all distribution shifts.
\item In this cyber-offensive query domain, attack success rates are extremely high.
\end{compactenum}

Firstly, \textbf{existing probe architectures can be significantly improved upon}. Prior work includes linear probing trained with mean aggregation (e.g. \citet{alain2016understanding, breakers}), attention probing \citep{kantamneni2025sparse, shabalin2025attention} or forms of max aggregation of trained linear probes \citep{cunningham2025cheap}. Our top performing probes: AlphaEvolve's MultiMax variation, and the Max of Rolling Mean Attention Probes, outperform these existing linear probe baselines. The Max of Rolling Mean Attention Probe with MultiMax aggregation outperforms all LLMs, and we strongly suspect that this is still far from optimal. 

Next, \textbf{no single method dominates across all distribution shifts}. As can be seen from \Cref{tabMainResults}, the optimal probe for a particular sub-distribution tends to differ from task-to-task. We suspect that this is due to using a wider set of distributions than prior work. Additionally, we speculate that the large variation in FPR and FNR values across techniques on a given domain is downstream of cyber-offensive capabilities being very similar to cyber-benign applications, and not significantly different from many coding related queries.

\begin{figure}[t]
    \centering
    \begin{subfigure}[t]{0.48\textwidth}
        \centering
        \includegraphics[width=\textwidth]{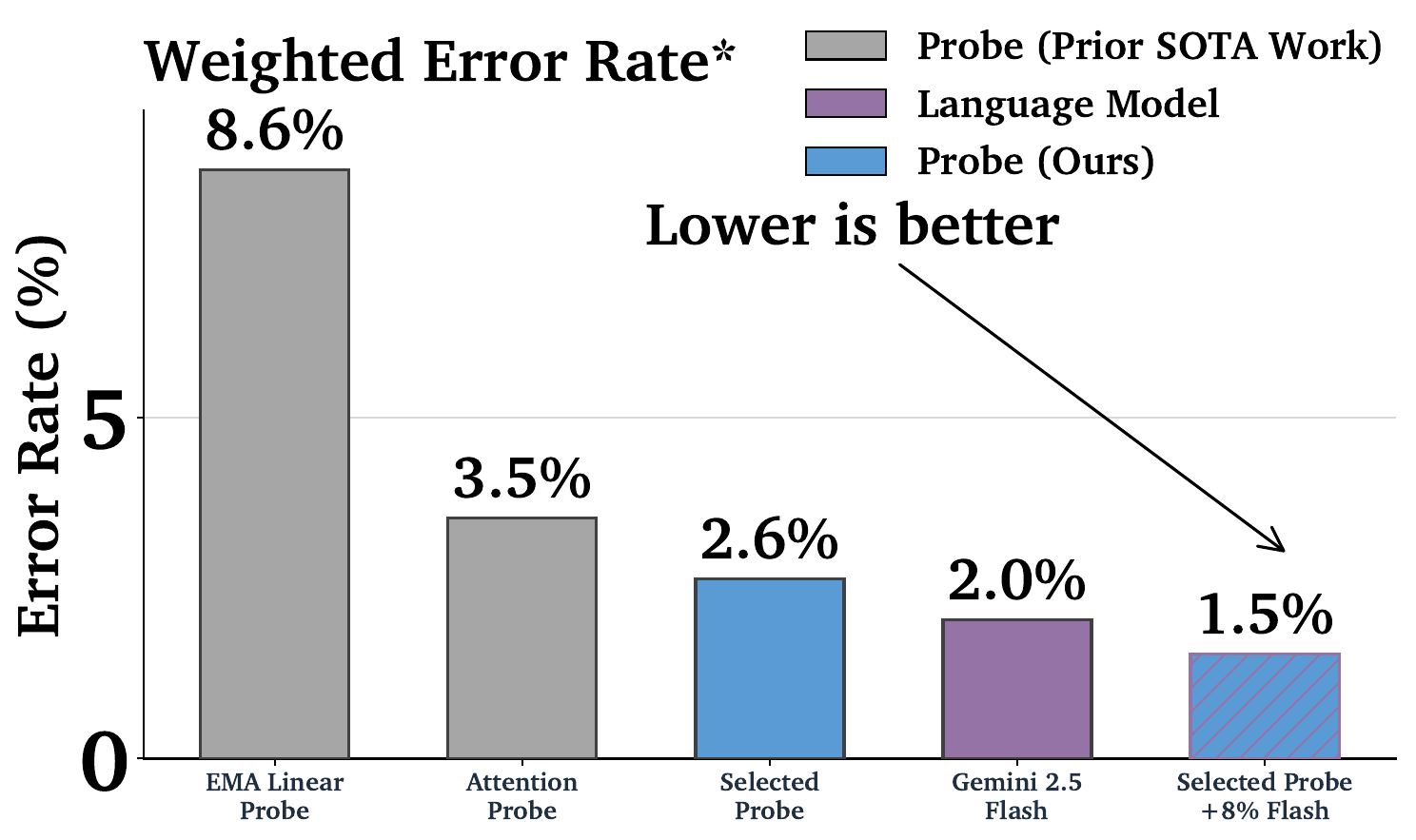}
        \caption{Test error comparison}
        \label{figTestLossComparison}
    \end{subfigure}
    \hfill
    \begin{subfigure}[t]{0.48\textwidth}
        \centering
        \includegraphics[width=\textwidth]{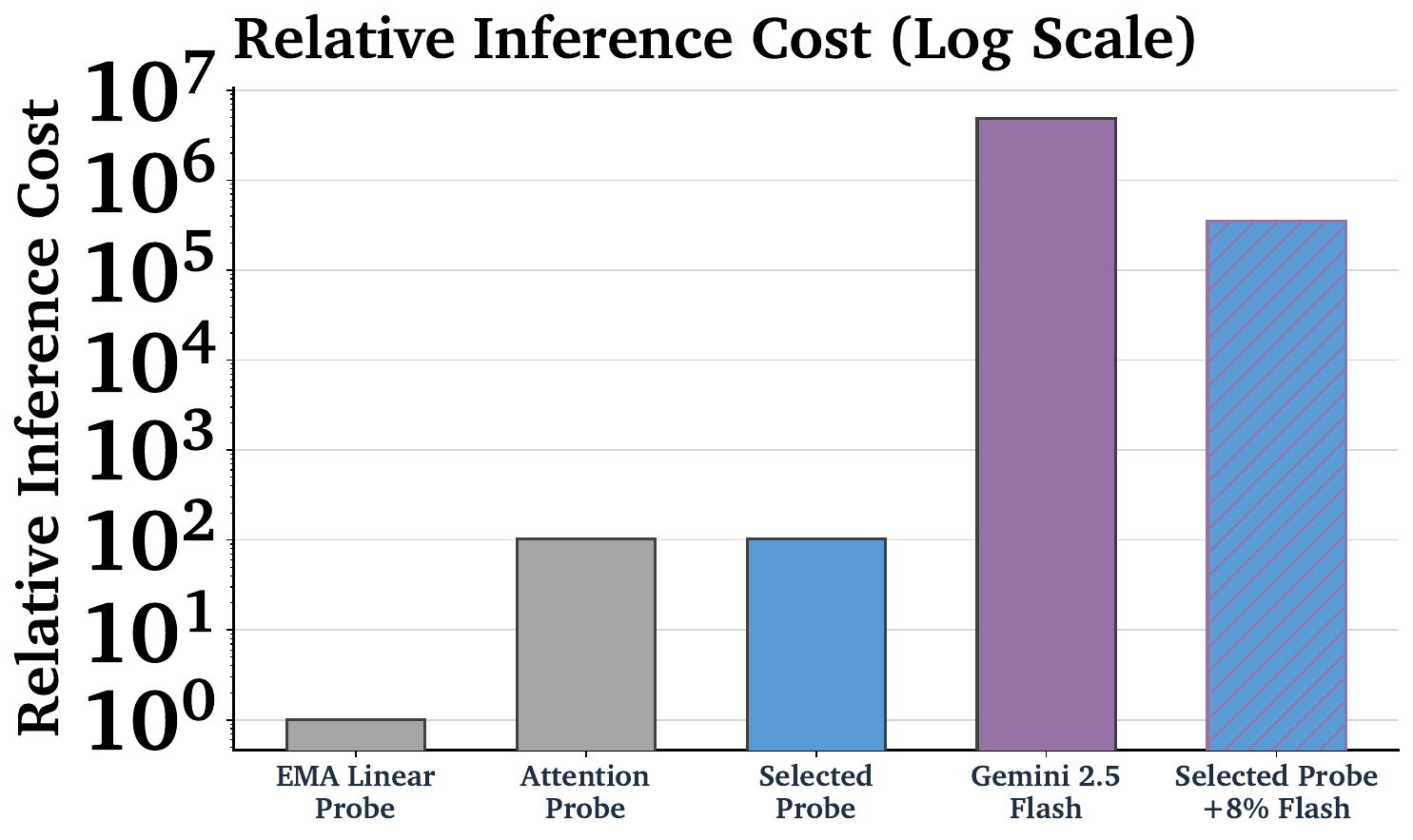}
        \caption{Cost comparison}
        \label{figCostComparison}
    \end{subfigure}
    
    \vspace{1em}
    
    \begin{subfigure}[t]{0.48\textwidth}
        \centering
        \includegraphics[width=\textwidth]{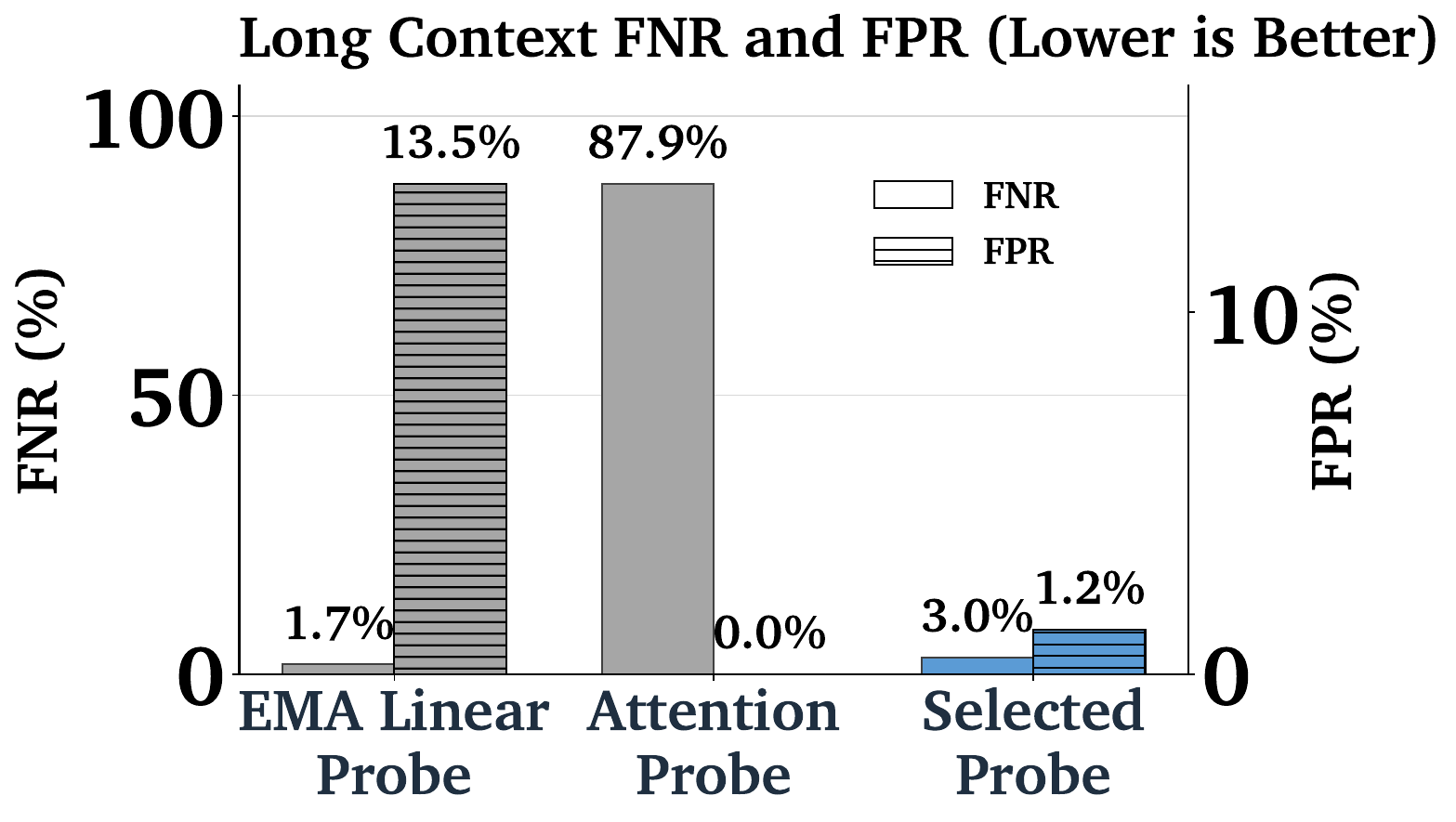}
        \caption{Long context FNR/FPR}
        \label{figLcFnrFprComparison}
    \end{subfigure}
    \hfill
    \begin{subfigure}[t]{0.48\textwidth}
        \centering
        \includegraphics[width=\textwidth]{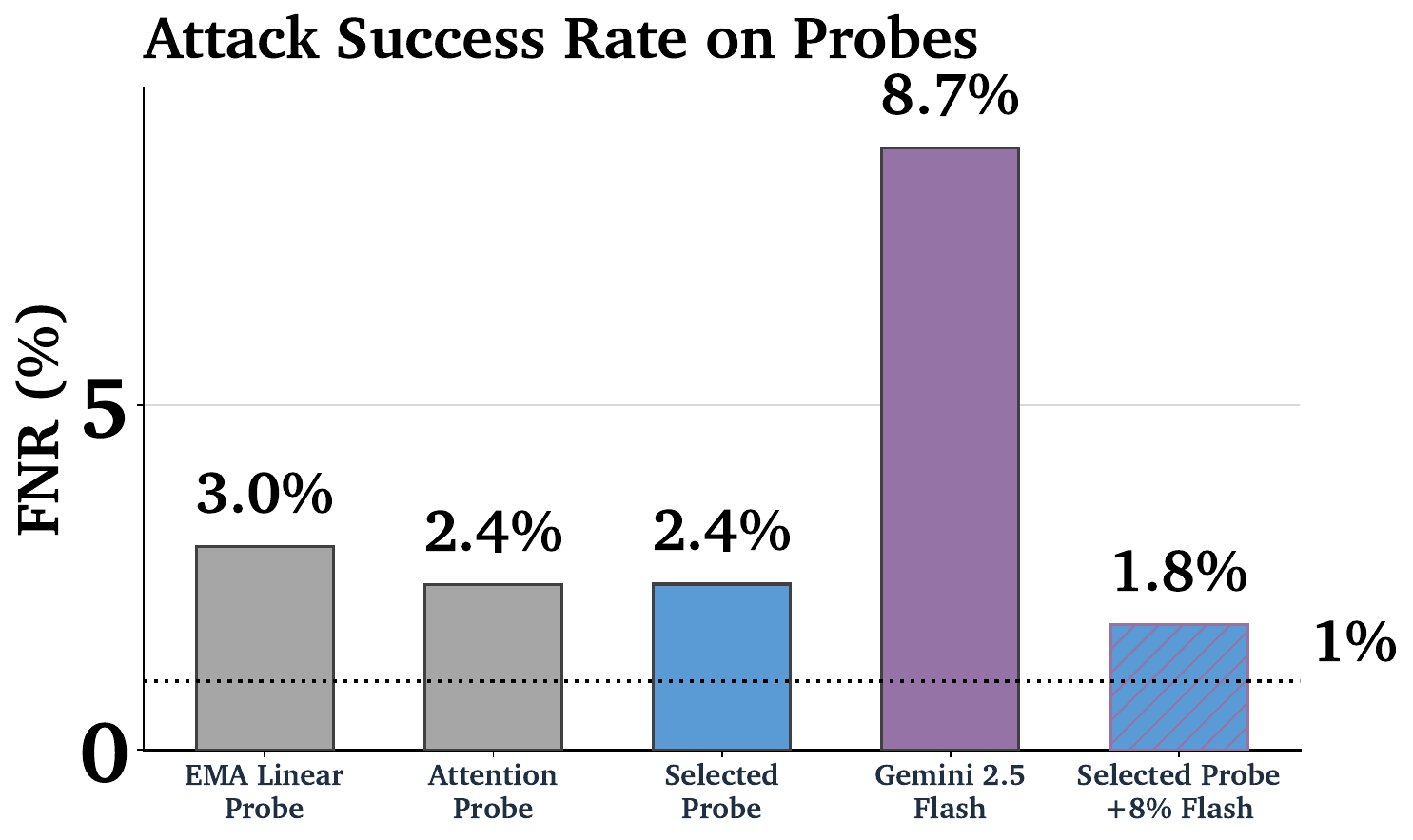}
        \caption{Pre-existing jailbreaks FNR}
        \label{figFnrArtComparison}
    \end{subfigure}
    \caption{Comparison of probe architectures across distribution shifts. (a) Test error across methods. (b) Probes achieve comparable performance to LLM classifiers at a fraction of the cost. (c) Long context generalization: AlphaEvolve probes maintain low FPR on long contexts. (d) Adversarial attacks remain challenging for all methods; all methods still leave attack success rate >1\%. Note: error bars are omitted from these plots see \Cref{figMainFigScatterCost,figCascadingClassifiers} for visualizations with (relatively wide) uncertainty estimates.\\
    \textit{*As in \Cref{figMainFigScatterCost} the threshold selection methodology is described in \Cref{subsecMetric}.}}
    \label{figDistributionsComparisons}
\end{figure}

Finally, \textbf{in this cyber-offensive query domain, jailbreak success rates are extremely high}. As gestured at in \Cref{figFnrArtComparison}, on the pre-existing jailbreaks dataset, all techniques have a vulnerability on at least 1\% of queries. This means that we are a long way from adversarial robustness in this domain. While we suspect that there is some dataset labelling error driving some low performance, we still believe that all of our probes and classifiers are far from saturating the task of detecting cyber-offensive queries. Note that this dataset consists of purely prompt based attacks generated from publicly accessible methods.

We now choose a \textbf{selected probe} to run further experiments with in this paper (e.g. \Cref{figMainFigScatterCost} and \Cref{figCascadingClassifiers}). \textbf{Our selected probe is the Max of Rolling Means Attention Probe}, which we find to statistically significantly outperform (according to bootstrap resampling) the baselines in \Cref{tabSignificance}. While the Max of Rolling Means Attention Probe, evaluated with a MultiMax aggregation function at inference time, has lower test error in \Cref{figMainFigScatterCost}, it also has far larger error bars, hence we do not focus on it.

\begin{table*}[t]
\centering
\caption{Main results across all evaluation datasets and classifiers. For methods with a configurable threshold $\tau$, $\tau$ is chosen to minimize \Cref{eqnTradeoff} on dataset validation splits. Better error rates are smaller; the best overall error rates in each column are bolded. Error bars are omitted; see \Cref{tabSignificance} for statistical significance testing.}
\label{tabMainResults}


\footnotesize
\setlength{\tabcolsep}{3pt}
\arrayrulecolor{gray!30}
\begin{tabular}{p{3cm}cccc|ccccc|c}
\toprule
\arrayrulecolor{black}
& \multicolumn{4}{c|}{\textbf{FPR} $\downarrow$} & \multicolumn{5}{c|}{\textbf{FNR} $\downarrow$} & \\
\cmidrule(lr){2-5} \cmidrule(lr){6-10}
Classifier & SC[OT] & SC[HN] & LC[RT] & MT[HN] & SC[A] & LC[A] & MT[A] & SC[J] & SC[ART] & \textbf{Test Error} $\downarrow$ \\
\midrule
\arrayrulecolor{gray!30}
Gemini 2.5 Flash & 0.3\% & 14.09\% & 0.7\% & 19.67\% & 8.58\% & 6.06\% & 13.51\% & 8.74\% & 20.65\% & \textbf{2.04\%} \\ \hline
Gemini 2.5 Pro & 0.2\% & 6.71\% & 0.2\% & 13.11\% & 16.01\% & 13.42\% & 20.85\% & 10.64\% & 33.87\% & 2.21\% \\ \hline
Attn (With LC Training) & 0.6\% & 7.04\% & 1.17\% & 4.92\% & 9.40\% & 1.34\% & 29.34\% & 3.33\% & 35.50\% & 2.38\% \\ \hline
Rolling Attn Probe, MultiMax Agg & 0.5\% & 12.75\% & 1.85\% & 11.48\% & \textbf{8.12\%} & 1.73\% & 22.01\% & 3.58\% & 22.97\% & 2.50\% \\ \hline
AlphaEvolve & \textbf{0.1\%} & 5.37\% & 1.72\% & 6.56\% & 9.74\% & 4.76\% & 32.05\% & 11.28\% & 27.61\% & 2.53\% \\ \hline
Rolling Attn Probe, Rolling Agg (Selected) & 0.7\% & 6.71\% & 1.23\% & 11.48\% & 8.58\% & 3.03\% & 26.25\% & 2.41\% & 42.46\% & 2.64\% \\ \hline
Rolling Attn Probe (20H), Rolling Agg & 0.5\% & 8.72\% & 1.0\% & 6.56\% & 11.37\% & 5.19\% & 32.05\% & 3.62\% & 50.58\% & 2.74\% \\ \hline
AlphaEvolve (Early Variant) & 0.2\% & 8.05\% & 1.11\% & \textbf{1.64\%} & 9.28\% & 7.79\% & 33.20\% & 42.28\% & 28.07\% & 2.87\% \\ \hline
Attn & 0.3\% & 10.07\% & \textbf{0.0\%} & 3.28\% & 9.98\% & 87.88\% & 35.91\% & 2.40\% & 43.39\% & 3.53\% \\ \hline
Gemini 2.5 Flash Lite & 0.7\% & 25.50\% & 0.2\% & 14.75\% & 11.37\% & 55.41\% & 27.41\% & 10.64\% & 41.76\% & 3.71\% \\ \hline
Rolling Attn Probe (20H), MultiMax Agg & 0.4\% & 9.40\% & 6.53\% & 24.59\% & 8.35\% & \textbf{0.0\%} & 15.44\% & 2.79\% & 22.74\% & 4.50\% \\ \hline
MultiMax Trained, MultiMax Agg & 0.3\% & \textbf{4.70\%} & 7.76\% & 3.28\% & 15.31\% & 0.4\% & 30.50\% & 5.17\% & 44.78\% & 5.35\% \\ \hline
Linear Probe Mean & 1.05\% & 11.41\% & 0.5\% & 8.20\% & 13.69\% & 99.13\% & 63.32\% & 70.48\% & 47.56\% & 6.18\% \\ \hline
Linear Probe Rolling Exp & 1.01\% & 16.78\% & 13.55\% & 75.41\% & 11.37\% & 1.73\% & 5.41\% & 2.95\% & 14.15\% & 8.63\% \\ \hline
MultiMax (Attn Trained) & 1.88\% & 17.45\% & 43.97\% & 70.49\% & 8.35\% & \textbf{0.0\%} & \textbf{2.70\%} & \textbf{0.8\%} & \textbf{3.94\%} & 21.93\% \\ \hline
MultiMax Trained (20H), MultiMax Agg & 1.12\% & 6.04\% & 82.14\% & 9.84\% & 9.51\% & \textbf{0.0\%} & 20.85\% & 7.61\% & 38.52\% & 38.14\% \\
\arrayrulecolor{black}
\bottomrule
\end{tabular}


\end{table*}


\subsubsection{Statistical Significance of Probe Comparisons}
\label{secStatisticalSignificance}

To rigorously validate our results, we compute the frequency with which one method outperforms another under a ``best-of-$k$'' selection procedure. In other words, to compare whether method A is better than method B, we are interested in the probability that when we train $k=100$ random seeds of each method and select the best by validation loss, method A achieves lower test loss than method B. We approximate this by using a bootstrap procedure over our seed sweeps, computing empirical PMFs (probability mass functions) for each method's test loss distribution under best-of-$k$ selection.

\Cref{tabSignificance} shows that our top methods---Rolling Attention Probe (with both Rolling and MultiMax aggregation) and AlphaEvolve---\textbf{statistically significantly outperform} all baseline methods. For comparisons against Linear Probe baselines, the bootstrap PMFs have \emph{zero overlap}: in every bootstrap sample, the top methods achieve lower test loss than the baselines. This validates that the performance gains we report are robust and not due to random seed selection.

\begin{table}[h]
\centering
\begin{tabular}{lcccc}
\toprule
 & \multicolumn{2}{c}{\textbf{Linear Baselines}} & \multicolumn{2}{c}{\textbf{Attention Baselines}} \\
\cmidrule(lr){2-3} \cmidrule(lr){4-5}
\textbf{Method} & \textbf{Linear Mean} & \textbf{Linear EMA} & \textbf{Attn} & \textbf{MultiMax} \\
\midrule
Rolling Attn, Rolling Agg (Selected Probe) & $1.0$ & $1.0$ & $> 0.996$ & $> 0.999$ \\
Rolling Attn, MultiMax Agg & $> 0.999$ & $> 0.999$ & $> 0.91$ & $> 0.99$ \\
AlphaEvolve & $1.0$ & $1.0$ & $> 0.96$ & $> 0.99$ \\
\bottomrule
\end{tabular}
\caption{Bootstrap frequency with which row methods outperform column methods under best-of-100-seeds selection. Values of $1.0$ indicate zero overlap between bootstrap PMFs. All selected probe comparisons exceed 0.95.}
\label{tabSignificance}
\end{table}

Among the top methods themselves, the differences are not statistically significant: the bootstrap frequency with which Rolling Attn (Rolling Agg) outperforms AlphaEvolve is only $0.51$, and the frequency with which it outperforms Rolling Attn (MultiMax Agg) is $0.48$, suggesting these architectures have converged to similar performance levels. This is visually illustrated in \Cref{figMainFigScatterCost}, where the ``Selected Probe'' and AlphaEvolve have overlapping error bars. This indicates we have identified a class of high-performing architectures, though determining the single ``best'' would require more seeds or a larger evaluation set.

\subsubsection{Training on Long-Context Data}
\label{subsubsecLCTraining}


In \Cref{tabMainResults} the ``Attn (LC Trained)'' baseline achieves a competitive test error ($\sim$2.38\% -- the architecture is described in \Cref{subsubsecAttention}). This confirms that attention probes have sufficient \textit{capacity} to handle long contexts if given enough data. However, these results rely on directly training on long-context data rather than generalizing from short-context data. 
This means that this generalization comes at a steep price: Even with large-scale accelerator clusters, the full dataset of long-context activations (each up to 1M tokens) cannot fit into TPU HBM. This necessitates implementing complex data loading pipelines to swap data between host memory and device memory, preventing the use of standard, efficient training loops. In our experiments, these overheads resulted in a \textbf{22$\times$ increase in training cost} compared to short-context training.

In contrast, our proposed \textbf{Max of Rolling Means Attention Probe} and \textbf{AlphaEvolve} architectures achieve similar performance (e.g., 2.53\% test error for AlphaEvolve compared to 2.38\% for Attn (LC Trained)) while being trained \textit{exclusively} on short-context data. This demonstrates that better architectural design can solve the long-context problem and achieve similar performance to directly training on long context data. 

We do note that attention probes may be optimal for domains without large deployment distribution shifts; in \Cref{appOtherDatasetsAndModels}, we find that attention probes are highly performant on open source datasets with the same train, test, and deployment distribution.

\subsection{Cascading Classifiers}
\label{secCascadingClassifiers}

Cascading classifiers combine a cheap probe with an expensive LLM classifier to achieve better cost-accuracy tradeoffs than either alone \citep{viola2001rapid, mckenzie2025detecting}. The probe first classifies each query; queries where the probe is uncertain are deferred to the expensive model. We consider two methods for computing the optimal deferral policy. Both methods set the parameters of the policy using the validation set.

\subsubsection{Warmup: Heuristic Band Method}

The heuristic band method uses a symmetric confidence band around the probe's decision threshold. Given a decision threshold $t$, we define a band $[t - \delta, t + \delta]$ and defer all queries with probe logits in this range to the expensive model. By varying $\delta$ from 0 (no deferral) to infinity (full deferral), we trace out a cost-accuracy curve. This method is simple but suboptimal: it treats all deferred queries symmetrically regardless of their individual costs.

\subsubsection{Generalization: Threshold-Randomization-Optimal Cascading}
\label{secOptimalFrontier}

The heuristic band method restricts us to symmetric deferral regions, but we can do better. In general, a cascading policy is defined by two thresholds $(t_0, t_1)$: classify as negative if the probe logit is below $t_0$, classify as positive if above $t_1$, and defer to the expensive model otherwise. Each choice of $(t_0, t_1)$ gives a point on a cost-accuracy plot on the validation data.

Since we have finite data, sweeping over all possible threshold pairs yields a discrete set of $(\text{Cost}, \text{Test Error})$ points. A natural question arises: what is the correct way to interpolate between these points to obtain a continuous Pareto frontier? We study and answer this question in detail in \cref{appOptimalCascadingPolicyAlgorithm}, and apply our results to the plot of the Pareto frontier in \cref{figCascadingClassifiers}.

\subsubsection{Cascading Results}

\begin{figure*}[t]
    \centering
    \includegraphics[width=\textwidth]{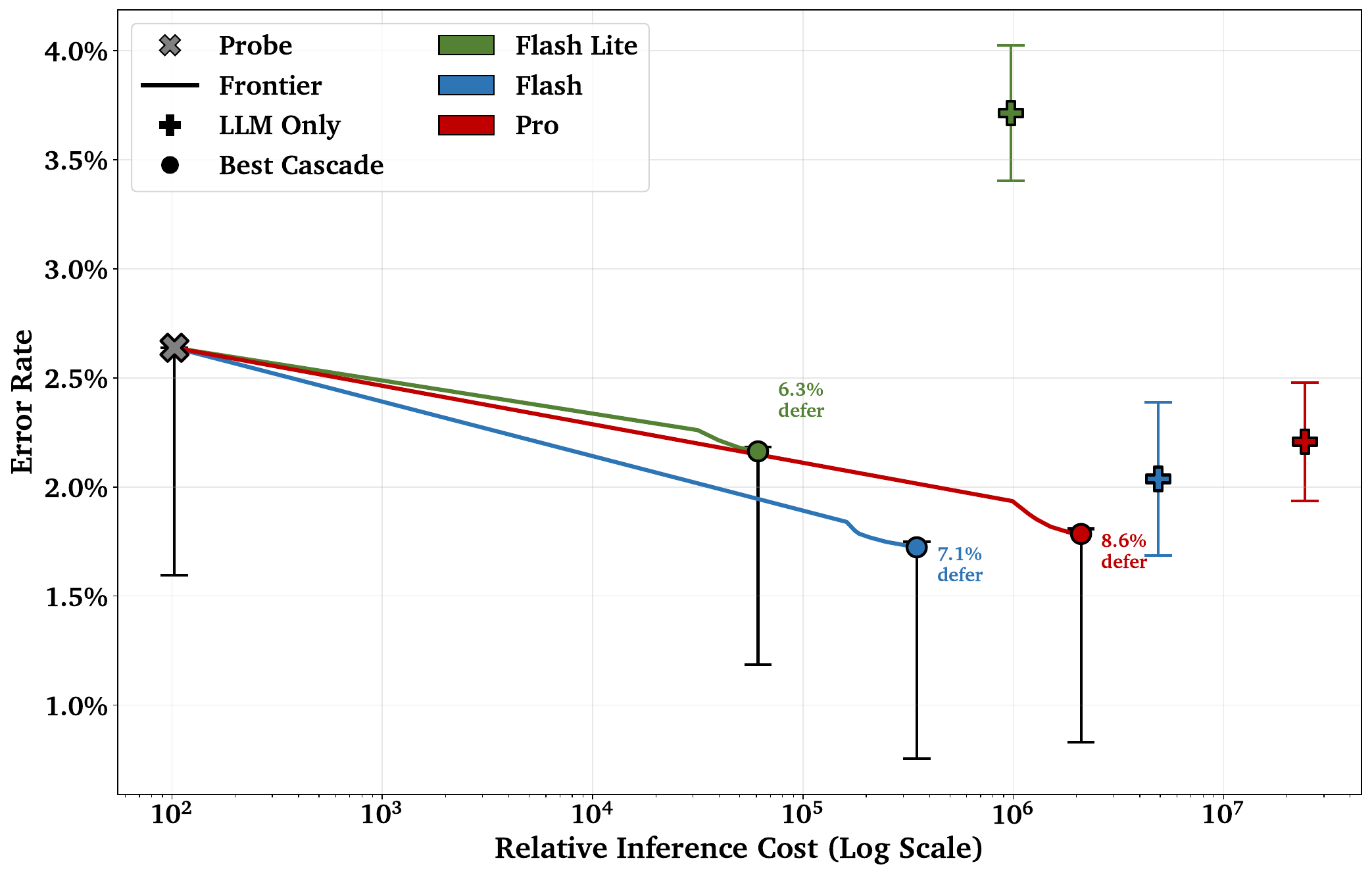}
    \caption{Cost plotted against test error (\Cref{eqnTradeoff}). The curves show the optimal frontier for cascading classifiers that combine cheap probes with expensive LLMs. Error bars (see \Cref{appErrorBars}) indicate substantial uncertainty: most operating points have overlapping confidence intervals, so the apparent ordering between methods should be interpreted cautiously. The cascade global minima (circles) represent optimal operating points for each LLM; further deferring to LLMs past these points increases the error rate.}
    \label{figCascadingClassifiers}
\end{figure*}

In \Cref{figCascadingClassifiers} we show the performance of cascading classifiers. Specifically, we find that Gemini 2.5 Pro and Gemini 2.5 Flash-Lite can be effectively combined with the optimal probe from \Cref{tabMainResults} in order to produce an even higher accuracy classifier than merely using the language models alone. The probe used on its own also lies on the Pareto frontier.

Notably, Gemini 2.5 Flash dominates Gemini 2.5 Pro across the entire cost-accuracy frontier. This is despite our cost estimation being generous to Pro: we ignore the cost of thinking tokens, treating them as free. In practice, Gemini 2.5 Pro often produces significantly more than the minimal 128 thinking tokens we set in the API, making the true cost gap even larger. This suggests that for cascading classifier deployments, Flash is the preferred expensive model. Note that all of these results are derived directly from sweeping over thresholds and then plotting the test data conclusion. However, since we're sweeping over one parameter only, we don't think this is a problematic use of the test data.

\subsection{Automating Safety Research with AlphaEvolve}
\label{secAlphaEvolveResults}

In this section, we describe our experience using AlphaEvolve to run an automatic evolutionary search for better probing architectures. 
AlphaEvolve \citep{novikov2025alphaevolve} is a system that optimizes algorithms by leveraging a large language model to iteratively mutate and refine programs. The system starts from a single seed program and then iteratively generates new programs using the high-performing programs as in-context examples for subsequent generation steps. High-performing programs are determined by their score on a user-defined automatic evaluation.

We give AlphaEvolve a simple base script that implements a naive attention probe. We then let AlphaEvolve optimize against a weighting of the validation error of datasets from \cref{tabDatasets} (excluding jailbreaks and long context overtriggering). We run a single large AlphaEvolve job that generates and evaluates approximately 2500 probing architectures. This process successfully closes approximately 50\% of the test error gap between the attention probe baseline and the best possible probe performance; see \Cref{figAlphaEvolve} for the the best generated architecture according to weighted validation error over time. 

The two AlphaEvolve architectures we compare to in the rest of this paper are chosen from this validation error frontier, one from relatively early in training and one from the very end. We show the corresponding pseudocode in \cref{subappAlphaEvolvePseudocode}. Notably, both of the chosen AlphaEvolve programs find MultiMax like solutions on their own. Indeed, the early AlphaEvolve program is just a MultiMax probe followed by a dense linear projection of the heads, while the program from the end of the AlphaEvolve run involves complicated gating and regularization.


We conclude with some speculation and observations about using AlphaEvolve for safety research; we expect our takeaways to be useful not just for AlphaEvolve, but also for similar automated evolutionary LLM-based program discovery systems.

\textbf{When to Use AlphaEvolve.} We think that automated program discovery methods are promising for safety research problems that can be framed as optimizing a single number or set of numbers. Indeed, we also successfully used AlphaEvolve to optimize the prompt for our adaptive jailbreak dataset (see \cref{secArtAnalysis}).

\textbf{Strange Late Effects in AlphaEvolve Generations.} We also observe some strange effects late in AlphaEvolve, where the best programs start to have strange comments describing hard to understand ideas like "Holographic Attention" or "frequency modulators"; we think this tendency is likely due to one of the prompts we used, which suggested that the model should ``Suggest a crazy idea of how we can improve our implementation, something that definitely nobody else would think of. Make it crazy with a capital C''. It may be important to balance increasing variance of model responses while preventing generation from going off of the rails. 

\textbf{High Level Recommendations On Using AlphaEvolve-like systems.} 
We believe that there are a number of important considerations that help to successfully use AlphaEvolve or similar systems. 
Chiefly, we recommend running multiple AlphaEvolve runs with increasing length and number of workers, tracking additional metrics that are not being expressly optimized by AlphaEvolve, and building environments that are as robust as possible to intense optimization pressure. Thus, a typical AlphaEvolve workflow for us looked something like: first designing an environment, running a small AlphaEvolve run on it, observing the results on held-out metrics, debugging why they lagged behind the optimized metrics, rerunning a larger AlphaEvolve run, debugging new compute bottlenecks in various parts of the pipeline (common for AlphaEvolve deployments), and so on. Thus, although AlphaEvolve can indeed greatly improve the performance of a seed program, there is a large overhead in building and iterating on a successful environment.

\textbf{Environmental Robustness: Evaluation Noise.} For the environment robustness problem in particular, there were two types of problems we found common. The first problem was the presence of noise in evaluation metrics, which is common in machine learning problems. For example, we believe that the large early drop in \cref{figAlphaEvolve} comes from climbing randomness, as some seeds are naturally better at the specific weighted tradeoff metric we specify. This is also reflected in our main results table in \cref{tabMainResults}: the AlphaEvolve programs do not improve over the seed Attention probe program on test metrics as much as one would expect from \cref{figAlphaEvolve}. We reduced variance by running multiple seeds in the AlphaEvolve evaluation function.\footnote{One interesting technique we experimented with to determine if we needed to invest more compute in reducing measurement noise was \textit{simulating} a simplified AlphaEvolve run, where we modeled each program run as a random process that had a distribution of improvements over the seed program and a distribution over measurement noise.} 

\textbf{Environmental Robustness: Reward Hacking.} The second type of environmental robustness problem we identified was (unintentional) reward hacking. For example, we found in early runs applying AlphaEvolve to our adaptive jailbreaking set, AlphaEvolve sometimes optimized for crashing inference servers by making the generated jailbreak extremely long, because a lack of a response was treated as a success, similar to \citet{tao}. We changed the metric to treat a lack of response as a failure, which mitigated this problem. In general, we found that surprising failures like this were common, motivating our above advice to iterate on environment design.


\section{Related Work}

\subsection{Misuse Mitigation}
The first line of defense for preventing language model misuse is aligning the model's behavior directly during training. Approaches in this vein include Reinforcement Learning from Human Feedback (RLHF) \citep{christiano2017deep} and Constitutional AI (RLAIF) \citep{bai2022constitutional}, which use feedback from humans or AI raters to teach the target model not to respond to harmful queries. However, these techniques are not yet sufficiently robust to prevent all misuse.

Recognizing that models may never be safe to use as standalone input/output black boxes, recent work augments language models with separate input output classifiers to flag potentially unsafe requests or responses \citep{naihin2023testing, inan2023llama, han2024wildguard, sharma2025constitutional}. The deployment of such monitors is now standard for frontier models, as detailed in frontier model system cards \citep{anthropic2025claude, deepthink}. The broader field of AI Control \citep{greenblatt2023ai} specifically focuses on augmenting untrusted language models into larger systems with components of language model, probe, or human oversight, although with more of a focus on model than user misbehavior. However, these text-based monitors can add significant latency and compute costs, so a promising alternative is \textit{activation probing}. 

\subsection{Activation Probing}
Activation probing originated as a diagnostic technique to determine what information is linearly decodable where from a frozen model's internal states \citep{alain2016understanding}. Recent work has adapted these diagnostic tools into active safety monitors \citep{zou2023representation}, demonstrating that probes can reliably detect high-stakes interactions \citep{mckenzie2025detecting} and latent deceptive intent \citep{goldowsky2025detecting}, often earlier and more efficiently than text-based classifiers. This efficiency is central to proposals for cheap misuse mitigation classifiers that reuse frozen internal features rather than processing the full output text with another language model \citep{cunningham2025cheap}. Contemporary research continues to refine probe architectures beyond simple linear layers, utilizing attention mechanisms \citep{shabalin2025attention}, exponential moving averages \citep{cunningham2025cheap}, or sparse autoencoders \citep{kantamneni2025sparse, tillman2025investigating} to improve performance.

\subsection{Cascading Classifiers}
Cascading classifiers\footnote{Sometimes called ``hierarchical classifiers'' or ``hierarchical monitors'', e.g. in \citet{mckenzie2025detecting}.} leverage a sequence of increasingly capable and more expensive models to optimize the trade-off between efficiency and accuracy (\citet{cascadingCertified, yue2024largelanguagemodelcascades}, though see \citet{mangal2022perilscascadingrobustclassifiers}). The foundational work by \citet{viola2001rapid} on rapid object detection demonstrated that a low cost layer could quickly reject the vast majority of negative examples, reserving expensive compute only for ambiguous cases. Recent work has adapted this paradigm to modern language model monitoring. \citet{greenblatt2023ai} propose monitoring architectures that combine trusted weak models, untrusted strong models, and trusted (but expensive) human oversight; via clever cascading design, they show that one can achieve much better performance than using any one source of oversight. Similarly, \citet{cunningham2025cheap, hua2025combiningcostconstrainedruntimemonitors, mckenzie2025detecting} demonstrate that activation probes can serve as a first line of defense for a more expensive language model classifier when detecting misuse, and \citet{cunningham2025cheap} specifically show that the resulting system achieves most of the performance of the language model while not costing significantly more than the extremely cheap activation probe. Finally, \citet{oldfield2025beyond} show that one can also apply this cascading classifier approach solely to a sequence of increasingly expensive activation probes (which correspond to additional polynomial terms in the probe's architecture), and \citet{youstra2025safeguardingllmfinetuningapis} show that probes are a stronger defence than LLMs for cipher attacks specifically.

\subsection{Distribution Shifts}
The failure of machine learning models to generalize when the test distribution differs from the training distribution is a classic and extensively studied problem \citep{blitzer2007biographies, torralba2011unbiased}. While large scale pretraining has generally improved out-of-distribution robustness compared to earlier architectures \citep{hendrycks2020pretrained}, modern language models remain vulnerable to many types of production distribution shifts (not least of which are adversarial examples; see the next related work section).
Notably for our paper, past work has observed that long context \citep{liu2023lost, hsieh2024ruler} and multi-turn data \citep{laban2025llms} are challenging distribution shifts for language models, so we should perhaps not be as surprised that activation probes—which directly operate on language model hidden states—struggle in these settings as well.

\subsection{Jailbreaking Models}
Adversarial examples—inputs explicitly designed to trigger model errors—are similarly a foundational and unsolved problem in machine learning \citep{szegedy2013intriguing}. In the context of language models, this manifests as "jailbreaking": crafting inputs that bypass safety training to elicit harmful outputs.

The discovery of these vulnerabilities has increasingly moved from manual engineering to automated search. \citet{zou2023universal} pioneered automated prompt finding with GCG, a white box gradient-based method for generating universal adversarial suffixes. Other approaches have leveraged LLMs as attackers: \citet{perez2022red} demonstrate the utility of using LLMs to generate diverse red teaming test cases, while PAIR \citep{chao2025jailbreaking} uses an iterative, adversary-versus-target LLM framework to refine attacks. Attacks have also emerged that exploit new model capabilities; notably, "many-shot" jailbreaking \citep{anil2024many} leverages extended context windows with hundreds of harmful demonstrations—a key distribution shift our probes are designed to withstand.

Benchmarks such as JailbreakBench \citep{chao2024jailbreakbench} and HarmBench \citep{mazeika2024harmbench} have been established to standardize jailbreak mitigation evaluation. However, \citet{nasr2025attacker} highlight that static defenses remain brittle, showing that adaptive attacks—where the adversary has knowledge of the defense—can bypass most existing guardrails.

\subsection{Automated Safety Research}
Recent advances have enabled language models to act as autonomous agents capable of conducting end-to-end scientific research; for example, \citet{lu2024ai} and \citet{gottweis2025towards} recently introduced frameworks for fully automated scientific discovery. Within the safety domain, such agents can accelerate manual red teaming and interpretability workflows; for example, MAIA \citep{shaham2024maia} uses a suite of tools to iteratively experiment on target models to identify failure modes and explain inner mechanisms.

Our work extends this trend by applying automated language model guided optimization directly to AI safety-relevant problems. \citet{novikov2025alphaevolve} propose AlphaEvolve, an evolutionary system that improves algorithms by prompting LLMs to make iterative changes to the best programs so far. We leverage AlphaEvolve in this paper to automate the search for well-generalizing probe architectures and to find more effective adversarial prompts in our automated red teaming system.

\section{Conclusion}
\label{secConclusion}

In this work, our \textbf{contributions} were: (i) showing that probes are an effective complement to black-box classifiers on cyber-misuse detection, (ii) showing that there is significant headroom in probe performance, with both architecture variations we discovered, and others that AlphaEvolve discovered, and (iii) highlighting that robustness to distribution shifts remains a difficult problem for probes (and LLM classifiers).

The \textbf{limitations} of our work include that we do not evaluate against probes using activations from every layer of the model as applied in concurrent work \citep{cunningham2026}, though potentially our findings and methodology (such as using AlphaEvolve) could be combined. Secondly, we focus entirely on monitoring inputs to language models, and while we study multi-turn inputs, we do not investigate whether any methods might classify that a language model rollout should be flagged partway through the rollout. Finally, a number of our results have large error bars, which makes it difficult for us to recommend one single probe architecture, for example. Future research could evaluate on several different domains beyond cyber misuse and academic datasets (\Cref{appOtherDatasetsAndModels}), to produce exact recommendations.

To close, we hope that our findings can both improve the quality of defenses used in frontier AI deployments at a given cost and also enable further research into the quality of activation probes. For example, we are excited about future work that builds on the open-weights and open-data research in \Cref{appOtherDatasetsAndModels}.

\section{Acknowledgements}
We are extremely grateful for the work of the \textit{Gemini Introspection} workstream at Google for their work adding probes to production Gemini deployments which we collaborated on. In particular Trilok Acharya, Wendy Kan, Felipe Tiengo Ferreira, Blaž Bratanič, Andrey Vlasov, Andre Fernandes, Tolu Delano, Weina Ge, Yuhan Fu, Chris Hsu, Felipe Tiengo Ferreira, Seiji Sakiyama, Thai Duong, Evan Feder, Julian Freedberg and Riccardo Patana all worked on this effort. We are additionally grateful to dataset and evaluation contributions from Ziyue Wang, Raluca Ada-Popa, Laurent Simon and Xerxes Dotiwalla, as well as Drew Proud, Tom Weingarten and the team behind the helpful-only Gemini variant we use, and finally Vikrant Varma, Tom Lieberum, Samuel Albanie, Irhum Shafkat and Kate Woolverton for fantastic foundational engineering tooling. Finally we are also grateful to Ziyue Wang for feedback on a draft of this paper, and grateful to Myriam Khan, Yotam Doron and Anca Dragan for publication help.

\section{Author Contributions}

JK made the first findings that MultiMax architectures could generalize to long contexts, and built foundational infrastructure that made all this work possible. JE ran all AlphaEvolve runs and wrote the first draft of the paper and edited it extensively, and additionally assisted with the autorater baseline experiments. ZW led the autorater baselines and ran the LC training experiments. BC built the seed program of the automated red-teaming infrastructure, and provided feedback on the paper. RS and NN provided advice throughout the project. AC proposed the initial project, ran all final probe evaluations except for LC Attn Probes, and wrote the final paper.

\printbibliography


%




\section*{Appendix}
\appendix
\crefalias{section}{appendix}
\crefalias{subsection}{appendix}

\section{Running our findings on other datasets and models}
\label{appOtherDatasetsAndModels}

To test the generality of our findings, we evaluate our probe architectures on a diverse set of classification tasks beyond cyber misuse detection. We use a subset of the binary classification datasets from \citet{kantamneni2025sparse}, selecting only datasets with at least 128 datapoints. The resulting 12 datasets span truthfulness (TruthfulQA), ethics (Commonsense Morality, Deontology, Justice, Virtue Ethics), news classification (Politics, Technology, Entertainment), as well as a simple text property detection task (Is Short Description). We also use the Gemma-2 9B \textit{base} model, like \citet{kantamneni2025sparse}.\footnote{E.g. \url{https://github.com/JoshEngels/SAE-Probes/blob/cec41e/ai_vs_humanmade_plot.py\#L21}}

\begin{figure}[tb]
    \centering
    \includegraphics[width=\columnwidth]{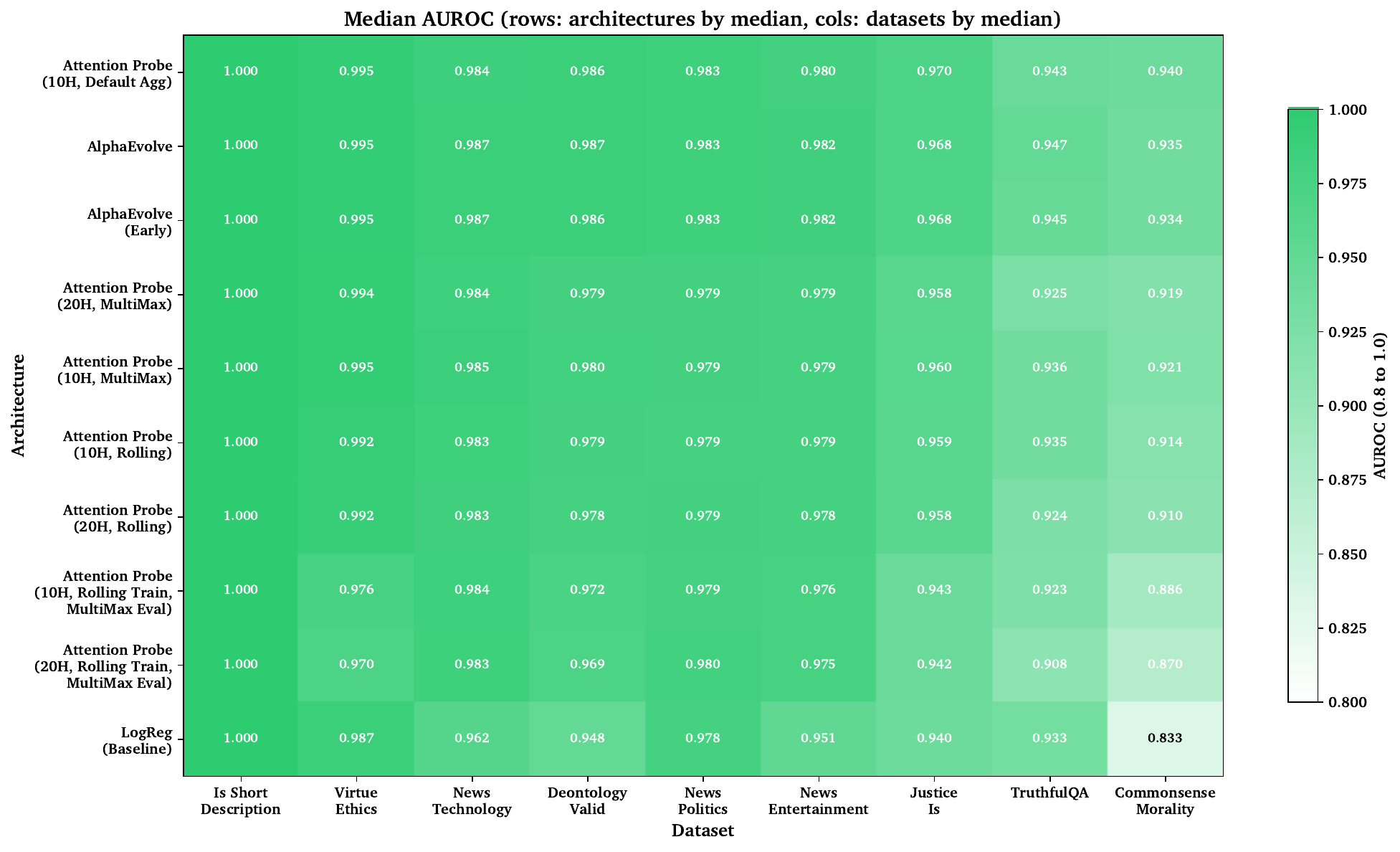}
    \caption{Median AUROC across 100 seeds for each architecture-dataset pair. Green indicates higher (better) AUROC values. Architectures (rows) and datasets (columns) are sorted by their overall median performance.}
    \label{figMultiDatasetMedian}
\end{figure}

\begin{figure}[tb]
    \centering
    \includegraphics[width=\columnwidth]{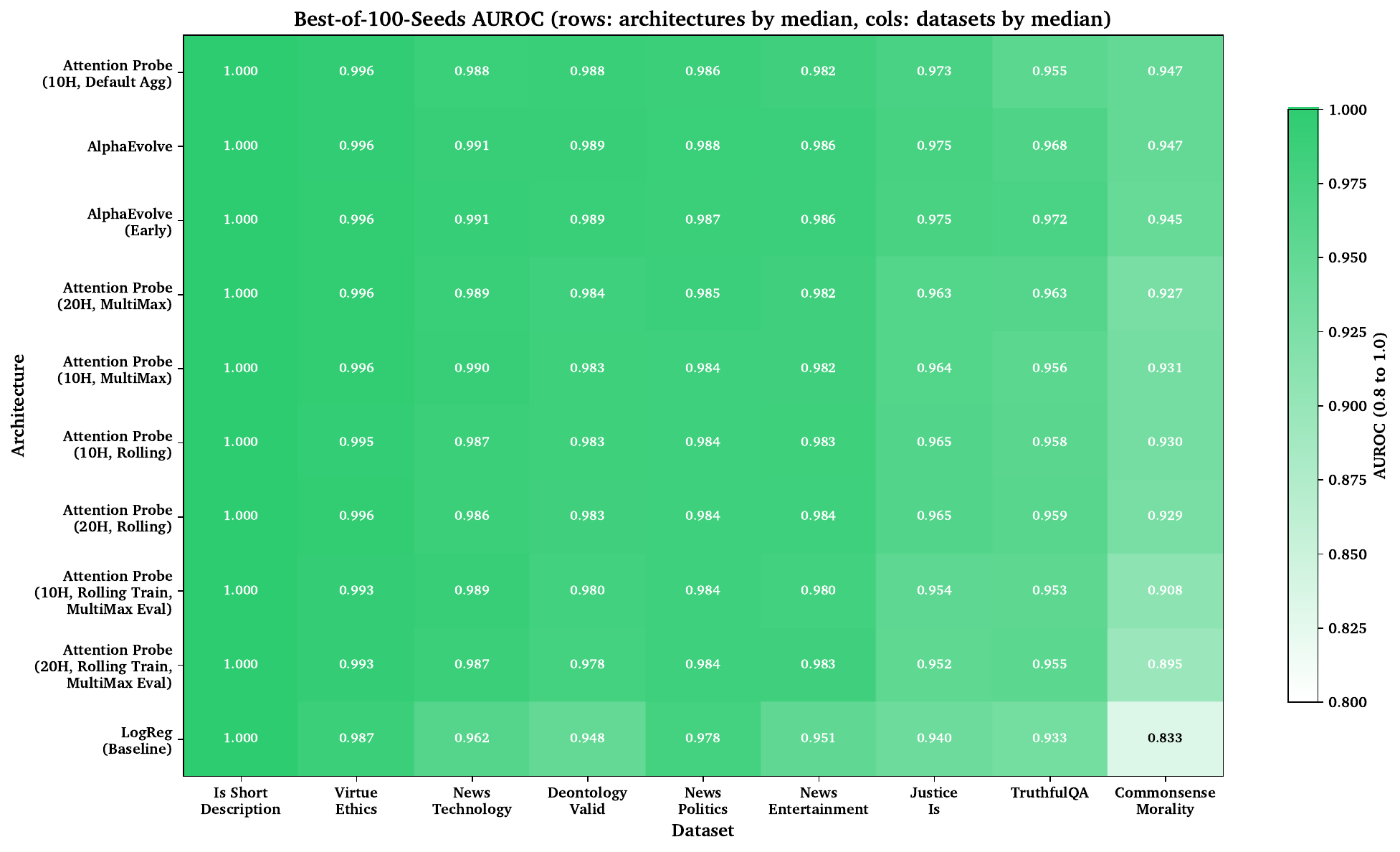}
    \caption{Best seed AUROC (out of 100 seeds) for each architecture-dataset pair. The ordering matches \Cref{figMultiDatasetMedian} to facilitate comparison.}
    \label{figMultiDatasetBest}
\end{figure}

\Cref{figMultiDatasetMedian,figMultiDatasetBest,tabMultiDatasetResults} show AUROC scores across all architectures and datasets. Each architecture is evaluated with 100 random seeds per dataset, allowing us to report both median performance and best-seed performance. As a baseline, we include the logistic regression results from \citet{kantamneni2025sparse}.

\paragraph{Architecture Ranking.} The AlphaEvolve architectures (early and final) achieve the highest median AUROC scores (0.975), followed closely by the Attention Probe with Default aggregation (median AUROC 0.975). The Attention Probe variants with MultiMax aggregation perform slightly worse (0.969), while the logistic regression baseline achieves median AUROC of 0.944. This ranking is consistent across most datasets, with AlphaEvolve outperforming the logistic regression baseline by an average of 3.1 percentage points in median AUROC.


\paragraph{Seed Variance.} Comparing median versus best-seed performance (\Cref{figMultiDatasetBest}), we observe that seed selection provides modest improvements of 0.5--1.5 percentage points depending on architecture.

\paragraph{Linear Probe Baseline.} Our experiments include a linear probe implementation trained using the same infrastructure as our other architectures. However, this linear probe achieves median AUROC of only 0.814 across datasets, substantially underperforming the logistic regression baseline from prior work (median AUROC 0.944). We attribute this gap to our training infrastructure being optimized for architectures with at least 10 or 100 times more parameters: the learning rate schedules, batch sizes, and regularization settings were tuned for attention-based probes, not linear probes. We therefore use the logistic regression baseline from prior work as the reference point for linear methods in our visualizations.

\paragraph{Comparison to Prior Work.} Several datasets overlap between our evaluation and Table 9 of \citet{kantamneni2025sparse}, which reports results on a sample of their full dataset collection; both works use Gemma 2 9B as the base model. On news classification, we observe slightly higher scores (0.98 vs 0.965). The largest difference appears on Commonsense Morality, where our attention probe achieves 0.94 compared to 0.86 in prior work. This gap may be explained by architectural differences: our attention probes include an MLP layer before the attention mechanism (\Cref{subsubsecAttention}), which provides additional representational capacity.

\begin{table}[tb]
\centering
\caption{Median AUROC by architecture and dataset. 
Dataset codes: IS=Is Short, VE=Virtue Ethics, 
NT/NP/NE=News Tech/Politics/Entertainment, 
DV=Deontology, JI=Justice, TQ=TruthfulQA, CM=Commonsense. 
Architecture codes: Def=Default aggregation, 
MM=MultiMax train/eval, Roll=Rolling train/eval, 
R$\to$MM=Rolling train with MultiMax eval.}
\label{tabMultiDatasetResults}
\footnotesize
\begin{tabular}{lcccccccccc}
\toprule
Arch & IS & VE & NT & DV & NP & NE & JI & TQ & CM & \textbf{Med} \\
\midrule
Attn-10H (Def) & 1.00 & 1.00 & 0.98 & 0.99 & 0.98 & 0.98 & 0.97 & 0.94 & 0.94 & \textbf{0.9834} \\
AE & 1.00 & 1.00 & 0.99 & 0.99 & 0.98 & 0.98 & 0.97 & 0.95 & 0.93 & \textbf{0.9833} \\
AE (Early) & 1.00 & 0.99 & 0.99 & 0.99 & 0.98 & 0.98 & 0.97 & 0.94 & 0.93 & \textbf{0.9830} \\
Attn-20H (MM) & 1.00 & 0.99 & 0.98 & 0.98 & 0.98 & 0.98 & 0.96 & 0.92 & 0.92 & \textbf{0.9791} \\
Attn-10H (MM) & 1.00 & 0.99 & 0.99 & 0.98 & 0.98 & 0.98 & 0.96 & 0.94 & 0.92 & \textbf{0.9791} \\
Attn-10H (Roll) & 1.00 & 0.99 & 0.98 & 0.98 & 0.98 & 0.98 & 0.96 & 0.93 & 0.91 & \textbf{0.9789} \\
Attn-20H (Roll) & 1.00 & 0.99 & 0.98 & 0.98 & 0.98 & 0.98 & 0.96 & 0.92 & 0.91 & \textbf{0.9782} \\
Attn-10H (R$\to$MM) & 1.00 & 0.98 & 0.98 & 0.97 & 0.98 & 0.98 & 0.94 & 0.92 & 0.89 & \textbf{0.9760} \\
Attn-20H (R$\to$MM) & 1.00 & 0.97 & 0.98 & 0.97 & 0.98 & 0.98 & 0.94 & 0.91 & 0.87 & \textbf{0.9701} \\
LogReg & 1.00 & 0.99 & 0.96 & 0.95 & 0.98 & 0.95 & 0.94 & 0.93 & 0.83 & \textbf{0.9513} \\
\midrule
Linear$^\dagger$ & 1.00 & 0.67 & 0.97 & 0.81 & 0.97 & 0.97 & 0.74 & 0.79 & 0.77 & \textbf{0.8145} \\
\bottomrule
\end{tabular}
\vspace{1mm}
\par\footnotesize{$^\dagger$Underperforms LogReg likely due to training infra optimized for larger probes.}
\end{table}

\section{Dataset Statistics}
\label{appDatasetStats}

Here we detail the specific sample counts and filtering logic for the datasets described in Table \ref{tabDatasets}.

\subsection{Evaluation and Validation Splits}

Table \ref{tabSampleCounts} provides the exact sample counts for the validation and test splits used to calculate the error metrics defined in Section \ref{secDatasetsAndMetrics}. 

\begin{table}[h]
    \centering
    \caption{Sample counts for Validation and Test splits across all datasets. The \textbf{Metric} column maps these datasets to the error variables defined in Table \ref{tabDatasets}.}
    \label{tabSampleCounts}
    \begin{tabular}{lcrrr}
        \toprule
        \textbf{Dataset Description} & \textbf{Metric} & \textbf{Val} & \textbf{Test} & \textbf{Total} \\
        \midrule
        \multicolumn{5}{l}{\textit{Benign / Random Traffic (False Positive evaluation)}} \\
        Short Context RT (Refusal/Response) & \FPRscot & 8,283 & 33,132 & 41,415 \\
        Long Context RT (Benign Code)       & \FPRlcrt & 202   & 812    & 1,014 \\
        \midrule
        \multicolumn{5}{l}{\textit{Attacks / Cyber (False Negative evaluation)}} \\
        Short Context Cyber                 & \FNRsca  & 483   & 918    & 1,401 \\
        Multi-Turn Cyber                    & \FNRmta  & 80    & 320    & 400 \\
        Long Context Cyber                  & \FNRlcsa & 429   & 1,718  & 2,147 \\
        Jailbreaks                          & \FNRscj  & 4,277 & 17,108 & 21,385 \\
        ART (Advanced Red Teaming)          & \FNRart  & 103   & 431    & 534 \\
        \bottomrule
    \end{tabular}
\end{table}

\subsection{Training Data Configuration}

Table \ref{tabTrainingData} details the composition of the training mixture. We apply heavy subsampling to the cyber-specific dataset and the Multi-Turn data to balance the dataset mixture.

\begin{table}[h]
    \centering
    \caption{Training dataset composition.}
    \label{tabTrainingData}
    \begin{tabular}{lr}
        \toprule
        \textbf{Dataset Description} & \textbf{Training Split Size} \\
        \midrule
        Short Context Attacks & 1,928 \\
        Short Context Overtriggering & 1,206 \\
        Multi-Turn Attacks & 41 \\
        \midrule
        \textbf{Total} & \textbf{3,175} \\
        \bottomrule
    \end{tabular}
\end{table}

\subsection{Long Context Data Filtering}

To ensure robust evaluation of Long Context (LC) capabilities without data leakage, strict filtering was applied to the \FNRlcsa dataset.

\begin{itemize}
    \item \textbf{Total LC stem fingerprints:} 2,147 raw rows.
    \item \textbf{Leakage Removal:} 1,711 LC rows matched the Short Context (SC) training data stems and were excluded from the LC evaluation to test true generalization.
    \item \textbf{Final Availability:} 436 LC rows were available for the `Clean' evaluation split.
\end{itemize}

\section{General Training Details}
\label{appTrainingDetails}

In this appendix, we list out various hyperparameters that we chose for our probe training runs. See also \Cref{appDatasetStats} where we describe the training, validation and eval data.

\begin{itemize}
\item \textbf{Optimizer}: AdamW
\item \textbf{Learning Rate}: $1 \times 10^{-4}$
\item \textbf{Weight Decay}: $3 \times 10^{-3}$
\item \textbf{Optimizer Momentum}: $\beta_1 = 0.9, \beta_2 = 0.999$
\item \textbf{Training Duration}: 1000 steps
\item \textbf{Batch Size}: Full batch gradient descent
\item \textbf{Source Layer}: Activations harvested from the layer halfway through the model
\item \textbf{MLP Transformation (\Cref{figClassifiers})}: Two layers of width 100
\item \textbf{Activation Function}: ReLU
\end{itemize}

\section{Seed Selection Details}
\label{appSeedSelection}

This appendix provides additional details on the seed selection analysis from \Cref{subsecSeeds}. 

\subsection{High-Variance Architectures}
\label{appSeedSelectionHighVariance}

\begin{figure}[tb]
    \centering
    \includegraphics[width=\columnwidth]{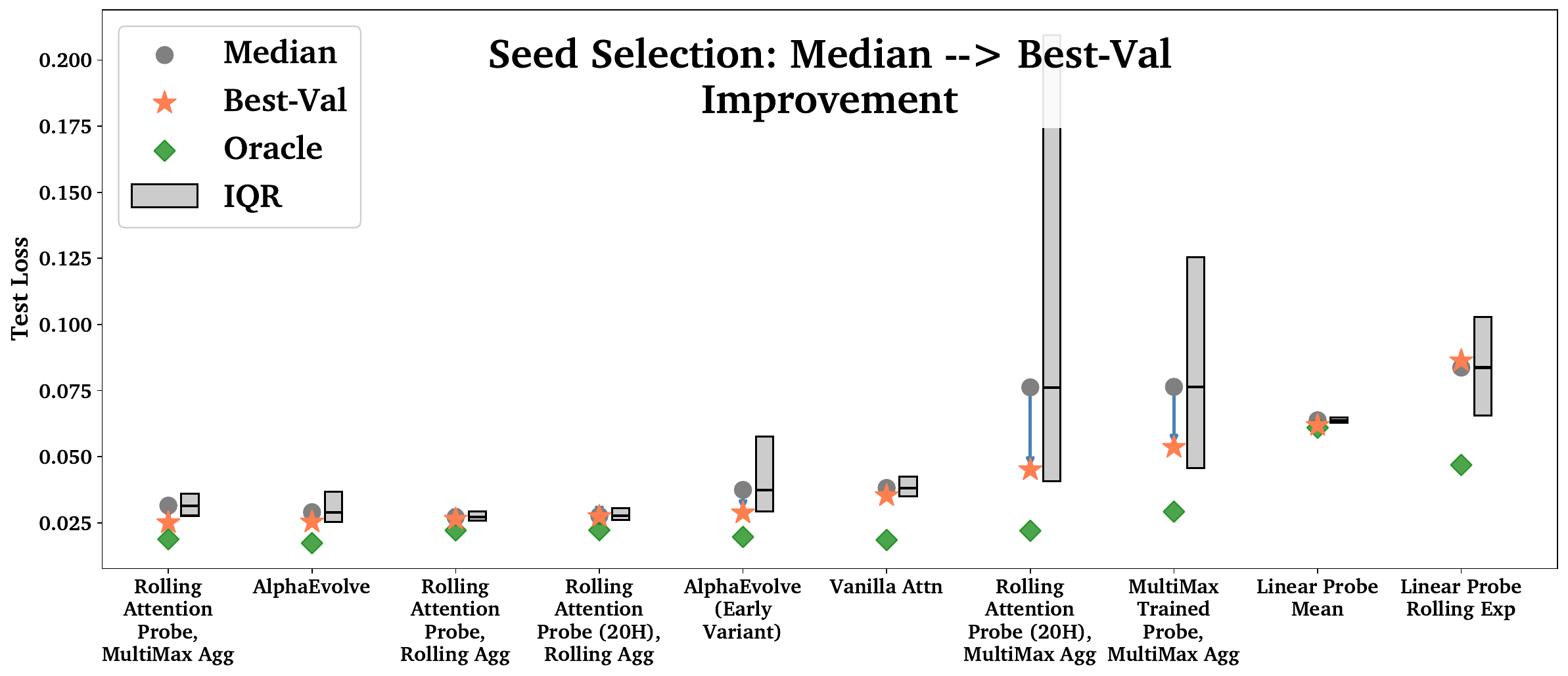}
    \caption{Effect of seed selection on test loss for high-variance architectures. Same format as \Cref{figSeedSelection}, but showing architectures with MultiMax aggregation that have large IQR across seeds.  
    }
    \label{figAppendixSeedSelection}
\end{figure}

\Cref{figAppendixSeedSelection} shows the seed selection analysis for architectures with MultiMax aggregation at evaluation time. These architectures exhibit substantially larger interquartile ranges (IQR) in test loss across seeds compared to those shown in \Cref{figSeedSelection}, which is why we present them separately to avoid distorting the main figure's scale.

\subsection{Raw Seed Selection Statistics}
\label{appSeedSelectionRaw}

\Cref{tabSeedSelectionRaw} presents the complete numerical data underlying \Cref{figSeedSelection} and \Cref{figAppendixSeedSelection}.

\begin{table}[tb]
    \centering
    \caption{Seed selection statistics across architectures. $\tau_{\text{med}}$, $\tau_{\text{best}}$, and $\tau_{\text{oracle}}$ are the test losses at the median, best-validation, and oracle (best-test) seeds respectively. $\Delta$ columns show the improvement from median to best-validation or oracle. A negative $\Delta_{\text{best}}$ indicates validation selection hurt test performance.}
    \label{tabSeedSelectionRaw}
    \small
    \begin{tabular}{lccccc}
        \toprule
        Architecture & $\tau_{\text{med}}$ & $\tau_{\text{best}}$ & $\tau_{\text{oracle}}$ & $\Delta_{\text{best}}$ & $\Delta_{\text{oracle}}$ \\
        \midrule
        Roll. Attn., MultiMax Agg & 0.031 & 0.025 & 0.019 & 0.006 & 0.013 \\
        AlphaEvolve & 0.029 & 0.025 & 0.017 & 0.004 & 0.012 \\
        Roll. Attn., Rolling Agg & 0.027 & 0.026 & 0.022 & 0.001 & 0.005 \\
        Roll. Attn. (20H), Rolling Agg & 0.028 & 0.027 & 0.022 & 0.000 & 0.005 \\
        AlphaEvolve (Early) & 0.037 & 0.029 & 0.020 & 0.009 & 0.018 \\
        Attn & 0.038 & 0.035 & 0.018 & 0.003 & 0.020 \\
        Roll. Attn. (20H), MultiMax Agg & 0.076 & 0.045 & 0.022 & 0.031 & 0.054 \\
        MultiMax Tr., MultiMax Agg & 0.076 & 0.054 & 0.029 & 0.023 & 0.047 \\
        \midrule
        Linear Probe Mean & 0.064 & 0.062 & 0.061 & 0.002 & 0.003 \\
        Linear Probe Rolling Exp & 0.084 & 0.086 & 0.047 & -0.003 & 0.037 \\
        \bottomrule
    \end{tabular}
\end{table}

\subsection{Seed Selection vs Architecture Choice}
\label{appSeedVsArch}

Consistent with the findings in \Cref{subsecSeeds}, we investigate the relative impact of seed selection versus architecture choice on these auxiliary datasets. To quantify the impact of seed selection, we compare the test AUROC of the best seed (oracle) to the median seed for each architecture. Across architectures, selecting the best seed improves test AUROC by 0.011 on average, though individual architectures vary from +0.007 (Attention Probe with Default aggregation) to +0.014 (Linear Probe).

In contrast, architecture choice provides a much larger gain. Comparing the median-seed AUROC, the best architecture (Attention Probe with Default aggregation) achieves 0.933 while the Linear Probe achieves only 0.824, a gap of 0.109 AUROC. This is approximately $10\times$ larger than the average seed selection gain, consistent with the main body's finding that architecture search should be prioritized over extensive seed tuning.

Interestingly, the AlphaEvolve architectures (AE1 and AE2) achieve nearly identical performance (0.930 median AUROC) and show higher seed variance (+0.012 gain) compared to the Attention Probe with Default aggregation (+0.007 gain). This suggests that more complex architectures may benefit more from careful seed selection, though the effect is still modest compared to architecture choice.

\section{Efficiently finding the Threshold-Randomization-Optimal cascading policy}
\label{appOptimalCascadingPolicyAlgorithm}
Recall that in general, a cascading policy is defined by two thresholds $(t_0, t_1)$: classify as negative if the probe logit is below $t_0$, classify as positive if above $t_1$, and defer to the expensive model otherwise. Each choice of $(t_0, t_1)$ gives a point on a cost-accuracy plot on the validation data. A natural question arises: what is the correct way to interpolate between these points to obtain a continuous Pareto frontier?

There are three possibilities: (1) a \emph{pessimistic} step function, where between operating points you are stuck at the accuracy of the lower-cost option; (2) \emph{linear} interpolation, where you can smoothly trade cost for accuracy; or (3) something \emph{better than linear}, where the curve bows upward (concave).

Linear interpolation (2) is achievable and optimal \emph{within the family of threshold-randomization policies}.\footnote{More sophisticated approaches exist: learned deferral policies can optimize routing based on input features rather than just probe classifier output \citep{mozannar2020consistent, madras2018predict}, and joint training can teach the cheap classifier to be ``usefully uncertain'' in the right places \citep{geifman2019selectivenet}. These methods may find operating points below our convex hull, but at the cost of additional training complexity. Our threshold-randomization approach has the advantage of being post-hoc: it can be applied to any pre-trained probe without retraining.} The key insight is that we can \emph{randomize} between two policies. Given two policies $A$ and $B$ on the frontier, we flip a biased coin and use policy $A$ with probability $p$ and policy $B$ with probability $1-p$. Since both expected cost and expected loss are linear in $p$, this traces out a straight line between the two points. In our setting, randomization is practical because we process many independent queries: rather than literally flipping coins, we can deterministically assign a fraction $p$ of queries to policy $A$. The resulting frontier is therefore the lower convex hull of all discrete operating points, which we compute efficiently in $O(N \log N)$ time (see \Cref{appOptimalCascadingPolicyAlgorithm} for algorithmic details). Importantly, when optimizing a single objective (such as minimizing error at any cost), the optimal policy always lies at a \emph{vertex} of the convex hull, meaning no randomization is required---the ``Selected Probe + 3\% Flash'' operating point in \Cref{figMainFigScatterCost} uses a single deterministic threshold pair (see \Cref{appVertexOptimality} for a formal argument).

We compute the optimal cascading frontier in $O(N \log N)$ time.\footnote{Note: the core details related to the Threshold-Randomization-Optimal policies and associated algorithms were suggested by Gemini 3.0 Pro DeepThink and other frontier models. But verified and extensively edited by \texttt{conmy@google.com} and frontier models.} The key insight is to frame the problem as \emph{greedy savings}: starting from a policy that defers everything to the LLM, we progressively ``buy'' samples to handle with the probe, always choosing the cheapest option first.

\paragraph{Setup.} For each sample $i$ (sorted by probe logit), we define three \emph{per-sample losses}: $\ell_-^{(i)}$ (loss if the probe predicts negative), $\ell_+^{(i)}$ (loss if the probe predicts positive), and $\ell_{\text{LLM}}^{(i)}$ (loss if we defer to the LLM). These are derived from the weighted loss function in \Cref{eqnTradeoff}. Note that we use ``loss'' here to distinguish from ``cost'' (the API expense of running the LLM).

\paragraph{The Greedy Savings Frame.} Consider the starting point: defer all $N$ samples to the LLM. This incurs maximum API cost but minimum error (the LLM's native error rate). From here, we can ``save'' LLM calls by having the probe handle samples instead. Each such move trades API cost for potential error (though note, the error might go down -- sometimes the probe is correct when the LLM is not!).

The crucial observation is that the left and right thresholds are \emph{independent}. We define two error functions:
\begin{itemize}
    \item $L(k)$: The added error if the probe classifies the $k$ lowest-logit samples as negative (instead of deferring them). This equals $\sum_{i=0}^{k-1} (\ell^{(i)} - \ell_{\text{LLM}}^{(i)})$.
    \item $R(k)$: The added error if the probe classifies the $k$ highest-logit samples as positive (instead of deferring them). This equals $\sum_{i=N-k}^{N-1} (\ell^{(i)} - \ell_{\text{LLM}}^{(i)})$.
\end{itemize}
Both functions map \emph{number of LLM calls saved} $\to$ \emph{test loss increase}. Since the left and right decisions don't interfere, the total added error for any policy is simply $L(k_L) + R(k_R)$ where $k_L + k_R$ is the total number of samples handled by the probe.

We now define the \emph{achievable} versions $\tilde{L}$ and $\tilde{R}$ as the lower convex hulls of $L$ and $R$ respectively. Points on the convex hull but not on the original curve are achievable via randomization between adjacent policies (see \Cref{secOptimalFrontier}).

The key insight is that the left and right decisions are \emph{independent}: choosing to handle $k_L$ samples on the left doesn't constrain which $k_R$ samples we handle on the right. The total added error is simply the sum $\tilde{L}(k_L) + \tilde{R}(k_R)$, and the total LLM calls saved is $k_L + k_R$. 

\begin{figure*}[t]
    \centering
    \includegraphics[width=\textwidth]{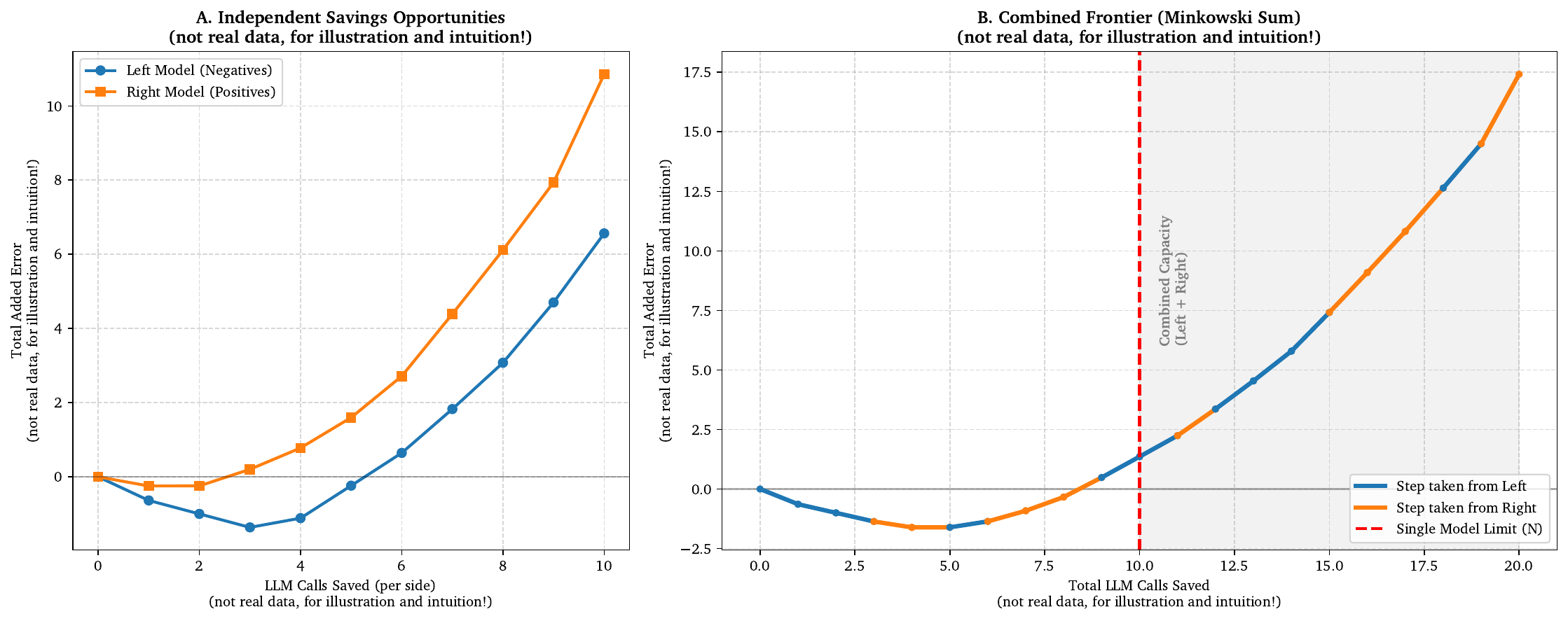}
    \caption{Illustration of the Minkowski sum algorithm for computing the optimal cascading frontier. \textbf{Left:} The two independent savings curves $\tilde{L}$ and $\tilde{R}$, representing the added error from having the probe handle samples from the left and right tails respectively. \textbf{Right:} The combined Pareto frontier of the Minkowski Sum of the two curves, constructed by merging edge segments sorted by slope. Segments are colored by their source (blue from left, orange from right). The vertical line marks the dataset size limit; beyond this point, savings are unachievable.}
    \label{figMinkowskiDemo}
\end{figure*}

Now we can visualize the set of points $(\text{LLM calls saved}, \text{Test error increase})$:\footnote{Of course, $k_L + k_R > N$ is not achievable since we only have $N$ samples total. We ignore this constraint for now, as working in this general form allows us to leverage existing algorithms for Minkowski sums. At the end, we simply restrict to the achievable region $k_L + k_R \leq N$; see \Cref{figMinkowskiDemo} for a visualization.}
\begin{equation}
    \mathcal{F} = \left\{ \left( k_L + k_R, \, \tilde{L}(k_L) + \tilde{R}(k_R) \right) : k_L \in [0, N], \, k_R \in [0, N] \right\}
    \label{eqnMinkowski}
\end{equation}
This is the \emph{Minkowski sum} of the two curves: $\mathcal{F} = \tilde{L} \oplus \tilde{R}$. we're combining all possible ways to ``spend'' our savings budget across the two independent markets. The Pareto frontier of this set traces out the optimal cost-error tradeoff (\Cref{figMinkowskiDemo}).

A beautiful result from computational geometry is that for convex polygonal curves, the Minkowski sum can be computed efficiently by simply merging the edge lists sorted by slope \citep{deberg2008computational}. Intuitively: at each step, we should accept whichever ``deal'' (left or right) offers the best marginal rate---the lowest added error per LLM call saved. Sorting by slope and greedily accepting the flattest edges first achieves exactly this.

\paragraph{The Algorithm.} We merge the edge lists of $\tilde{L}$ and $\tilde{R}$ sorted by slope. Walking this merged list from flattest to steepest slope traces out the Pareto frontier: at each step, we accept the cheapest available ``deal'' regardless of which side it comes from. We stop when we have saved all $N$ LLM calls (i.e., the probe handles everything).

The convex hull computation is $O(N)$ after sorting, and merging two sorted lists is $O(N)$. The initial sort by probe logit dominates at $O(N \log N)$.

Note that ``LLM calls saved'' is the flipped version of ``cost'' used in \Cref{figCascadingClassifiers}. This is why the Pareto frontier in \Cref{figMinkowskiDemo} curves upward (more savings $\to$ more error), while in \Cref{figCascadingClassifiers} it curves downward (more cost $\to$ less error).

\subsection{Vertex Optimality: Why No Randomization is Needed At The Optimal Low Error Point}
\label{appVertexOptimality}

A natural question arises: if randomization between policies is allowed, does the optimal policy (the one minimizing error) require randomization? We show that the answer is \emph{no}---the optimal policy always lies at a vertex of the convex hull, corresponding to a single deterministic threshold pair.

\paragraph{Geometric Argument.} Consider optimizing any linear objective over a convex set. The level sets of a linear function are hyperplanes, and the minimum over a convex polytope is always achieved at a vertex (or along an edge, in degenerate cases). In our setting:
\begin{itemize}
    \item The set of achievable (Cost, Error) pairs is the convex hull of discrete operating points---a convex polygon.
    \item Minimizing error alone corresponds to a horizontal objective (minimize the $y$-coordinate).
    \item More generally, any weighted combination of cost and error is linear and thus minimized at a vertex.
\end{itemize}

The vertices of our convex hull correspond exactly to non-randomized policies: each vertex is a specific threshold pair $(t_0, t_1)$. Interior points and edges require randomization between policies.

\paragraph{Application to \Cref{figMainFigScatterCost}.} The ``Selected Probe + 8\% Flash'' operating point in \Cref{figMainFigScatterCost} is the vertex that minimizes error. This corresponds to a single deterministic cascading policy with specific thresholds $(t_0, t_1)$, where approximately 3\% of queries (those with probe logits between $t_0$ and $t_1$) are deferred to the LLM. No randomization is needed to achieve this optimal operating point.

This is convenient in our setting, where we seek to minimize error: deterministic policies are easier to implement, debug, and reason about than randomized policies. If we instead targeted a specific cost budget, randomization might be required to achieve points on the frontier edges. Additionally, for misuse defense applications, randomized policies may be undesirable from a security perspective: \citet{hughes2024bestofnjailbreaking} show that attackers can exploit randomization by repeatedly querying until they receive a favorable response (``Best-of-N jailbreaking''). While our defenses already face challenges from adaptive attacks (\Cref{figFnrArtComparison}), deterministic policies avoid introducing this additional attack surface.

\section{AlphaEvolve}
\label{appAlphaEvolve}

\subsection{Further Details on Setup}
\label{subappAlphaEvolveSetup}

The \textbf{training data} that AlphaEvolve trains on is 70\% of the short-context cyber attack data (\Cref{tabDatasets}) combined with 70\% of the short-context overtriggering examples. The remaining 30\% of each is used for validation threshold selection (see below). Long-context data is only used for validation error computation, not for training.

The \textbf{validation error} which we hillclimb with AlphaEvolve is the maximum of weighted error rates across different error types: 200$\times$ for overtriggering FPR, 40$\times$ for long-context FPR, 6.7$\times$ for cyber FPR, and 4$\times$ for both cyber FNR and long-context FNR. Note that this weighting differs from the 5:2:50 weighting used in \Cref{eqnTradeoff} because we ran AlphaEvolve early in this research project before settling on the final metric.

On net this means that inside the inner training loop, the only data used is training data, according to \Cref{tabDatasets}. In validation, which is used in AlphaEvolve as feedback to the LLM generating programs, we use validation data, as well as some long-context data, which in the main body is only used for evaluation. We retrained the AlphaEvolve architectures with the final training mixture (e.g. with Multi-Turn data) and reported those numbers in the same body. Additionally \Cref{appOtherDatasetsAndModels} reports the same AlphaEvolve architecture on a different model and with different data, where it performs strongly, which suggests to us that we didn't significantly contaminate our results. Notably, in \Cref{figAlphaEvolve}, where we find that AlphaEvolve is the second strongest architecture after the Attention Probe, and the Attention Probe would not generalize to long context prompts without specific training (see the main paper body). We speculate that rerunning AlphaEvolve with the same training mixture it is to be trained with would lead to even stronger performance, but we haven't verified this. 

\subsection{Alpha Evolve Probe Pseudocode}
\label{subappAlphaEvolvePseudocode}

Early in the AlphaEvolve run (\Cref{figAlphaEvolve}), we found that AlphaEvolve rediscovered the MultiMax architecture (although note that we provided it with an Attention Probe as a base). AlphaEvolve also added a learnable linear feature map to weight each of the head outputs rather than merely sum them. This makes this probe the only such variant that we study in this paper which has a step 6) in \Cref{figClassifiers}, \textit{not} merely a sum. This architecture (MultiMax plus a linear weighting of heads) is the early in training AlphaEvolve program described in our main results.

We present pseudocode for the final best-performing AlphaEvolve-discovered architecture in \Cref{algGatedBipolar}. The architecture combines several innovations: gated projections using Softplus activations, a ``bipolar'' pooling strategy that takes both the max and negated min across the sequence dimension (generalizing MultiMax), and regularization via L1 penalties and orthogonality constraints on the projection weights.

\begin{algorithm}[t]
    \SetAlgoLined
    \KwIn{Dataset of activations $\{\vX_i\}$ and labels $\{\ell_i\}$, learning rate schedule $\eta_t$, reg weights $\lambda_1, \lambda_{ortho}$}
    \KwResult{Trained probe parameters $\theta = \{W_{enc}, W_{proj}, W_{gate}, W_{out}, \dots\}$}
    \While{not converged}{
        
        $H \leftarrow \text{MLP}(\text{LayerNorm}(\{\vX_i\}))$\;
        
        \tcp{Compute gates}
        $V \leftarrow (W_{proj} H) \odot \text{Softplus}(W_{gate} H)$\;

        \tcp{Compute max and min across sequence dim (generalization of MultiMax)}
        $h_{pool} \leftarrow \text{Concat}(\max_{t} V_t, -\min_{t} V_t)$\;
        
        $y \leftarrow W_{out} h_{pool}$\;
        
        \tcp{Compute loss with L1 and orthogonality penalties}
        $\mathcal{L}_{BCE} \leftarrow \text{BinaryCrossEntropy}(\sigma(y), \{\ell_i\})$\;
        
        $\mathcal{L}_{reg} \leftarrow \lambda_1 \sum_{W \in \theta} ||W||_1 + \lambda_{ortho} ||W_{proj}^T W_{proj} - I||_F^2$\;
        
        $\theta \leftarrow \text{AdamW}(\theta, \nabla_\theta(\mathcal{L}_{BCE} + \mathcal{L}_{reg}), \eta_t)$\;
        
    }
    \caption{Pseudocode for Best-Performing Alpha Evolve Probe Architecture.}
    \label{algGatedBipolar}
\end{algorithm}

\subsection{Training Curve}
\begin{figure}[t]
    \centering
    \includegraphics[width=\columnwidth]{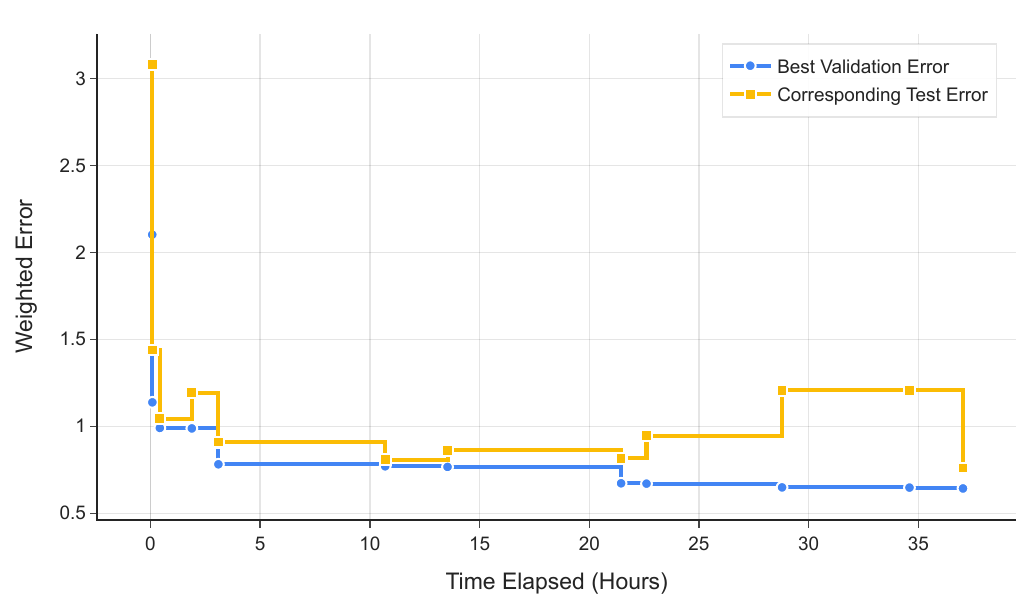}
    \caption{Weighted validation and test errors for best probe architecture over the course of the AlphaEvolve run. The process successfully closed approximately 50\% of the test error gap between the attention probe baseline and perfect probe performance.}
    \label{figAlphaEvolve}
\end{figure}

\Cref{figAlphaEvolve} shows the training curve for the AlphaEvolve run.

\section{Prompt Optimization Experiments}
\label{appHillclimbingPrompt}

\begin{figure}[t]
	\centering
	\includegraphics[width=\columnwidth]{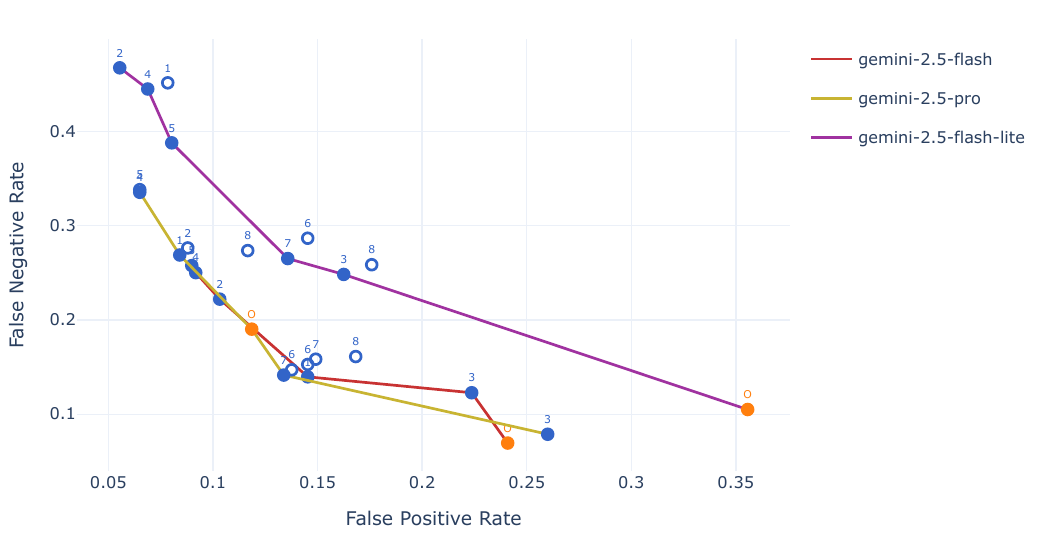}
	\caption{
    Different prompts trade off on \FPRscot~and \FNRsca (results are shown on the training set of these datasets). The prompts that we try are labeled by the order we created them; the original prompt is in orange. Prompts that lie along the pareto frontier are solid.
    }
	\label{figPromptComparison}
\end{figure}

In addition to our original cyber misuse detection prompt provided by colleagues, we manually create eight additional prompts with varying amount of detail and instruction. For some of these prompts, we use mild manual prompt optimization by identifying examples where existing classifiers make mistakes and adding instructions that aim to correct these mistakes, while we greatly simplify some of the other prompts. We use the test sets of the short context overtriggering and short context attack datasets for evaluating the prompts.  Our main takeaway from this section is that we have evidence that this original provided prompt is reasonably optimized.

The nine total prompts form an FPR vs. FNR curve for each model, shown on the left side of \cref{figPromptComparison}. While different prompts can indeed shift along the curve, no prompts are better than the original prompt on both FPR and FNR. Furthermore, the curve is approximately linear, so e.g. different prompts seem to be able to shift a constant number of negatives for positives. It therefore seems that different prompts do not significantly improve upon each other, so we stick with our original prompts.

\section{Automated Red Teaming Analysis}
\label{secArtAnalysis}

\begin{figure}[t]
	\centering
	\includegraphics[width=\columnwidth]{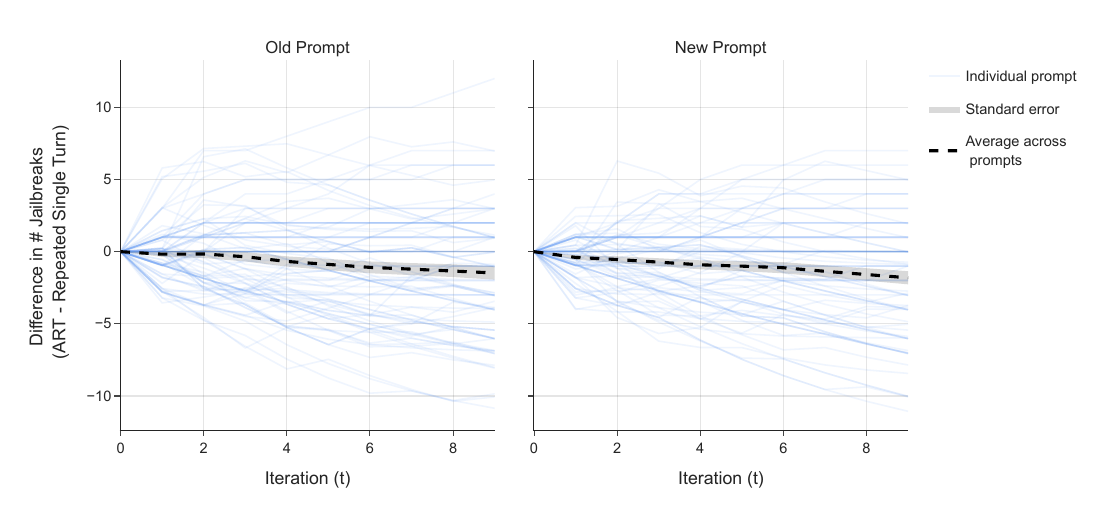}
	\caption{Multiple adaptive red teaming steps do not help jailbreak success compared to simply repeating the first step.}
	\label{figJailbreakComparison}
\end{figure}

We developed an iterative adaptive attack framework where an attacker model is given $T$ turns to jailbreak the target model. At each turn, the attacker receives its history—including whether previous attempts (1) bypassed the monitor and (2) jailbroke the target model (verified by an autorater) -- allowing it to learn from failures. We evaluated this system on 100 random queries from the short context attacks test set, running $N=20$ seeds per query.

In initial experiments with $T = 10$, we found that adaptive turns did not outperform independent repetitions of the first turn. Formally, consider a prompt $S$ with $N$ seeds. Let $M$ be the number of successful jailbreaks observed in the first turn. The expected success probability (at least one jailbreak) if we repeated the first step independently $T$ times is $\mathcal{P}_{\text{indep}}(S, T) = 1 - (1 - \frac{M}{N})^T$. We compare this to the observed success rate of the adaptive procedure on the same seeds, $\mathcal{P}_{\text{adapt}}(S, T)$. Across all 100 prompts $S_i$, we found that $\sum_{i} \mathcal{P}_{\text{indep}}(S_i, T) > \sum_i \mathcal{P}_{\text{adapt}}(S_i, T)$ for all $T > 1$. Repeating the first iteration was superior to the full iterative procedure (see \Cref{figJailbreakComparison}). We note this does not contradict \cite{chao2025jailbreaking}, as they do not compare to this baseline. Consequently, for the main results, we simplified our setup to $T = 1$.

We subsequently applied AlphaEvolve to optimize our attacker prompts (not released), improving the single-turn jailbreak success rate from 20\% to 26\%. We utilize this improved prompt for our main results. However, even with these optimized prompts, the inequality $\sum_{i} \mathcal{P}_{\text{indep}}(S_i, T) > \sum_i \mathcal{P}_{\text{adapt}}(S_i, T)$ held for all $T > 1$.

Finally, we tested a "cross-pollination" variant where the attacker received $k=3$ random successful and failed attacks from other seeds at prior steps. While this intervention achieved $\sum_{i} \mathcal{P}_{\text{indep}}(S_i, T) < \sum_{i} \mathcal{P}_{\text{adapt}}(S_i, T)$, the margin was negligible. Given the significant complexity and computational cost increase, we leave further evaluation of such systems to future work.

\subsection{Definition of ``Success'' in Adaptive Red Teaming}
\label{appArtSuccessDefinition}

A natural concern with our ART dataset construction is that the narrative rewriting process might inadvertently transform a malicious prompt into a benign one. If this happened, ``successful'' jailbreaks would simply reflect easier queries rather than genuine vulnerabilities.

We guard against this by defining ``success'' using an autorater that evaluates the \textit{model's response}, not just whether the monitor was bypassed. Specifically, an attack is considered successful if and only if:
\begin{enumerate}
    \item The attack bypasses the monitor (i.e., is not flagged), \textbf{and}
    \item An autorater judges the model's response to be harmful.
\end{enumerate}

The autorater compares the model's response to a baseline response from ``adversarial Gemini'' (Gemini without safety training, sometimes called a ``helpful-only'' variant of the model. This model is not publicly released) given the \textit{original, un-rewritten} harmful query. This ensures that both responses are evaluated against the same underlying malicious intent. An attack is only counted as successful if the response to the rewritten prompt is at least as harmful as the baseline.

This design holds query difficulty constant: if the rewriting process neutralizes the harmful intent, the model will produce a benign response that fails the autorater check, and the attack will not be counted as successful. Our reported success rates (e.g., improvement from 20\% to 26\% via AlphaEvolve optimization) therefore reflect genuine jailbreak rates on queries of consistent difficulty.



\section{Infrastructure Recommendations}
\label{appInfra}

We recommend storing activations on disk when training single-layer probes. We find that this makes iterating on probe architecture (e.g. manually or with AlphaEvolve) much faster. This advice contrasts with \citet{cunningham2026}, who recommend generating activations from the base LLM during probe training. We believe that this difference is due to our probe training using activations from a \textit{single} layer (so the storage cost and loading time is only proportional to the dataset size and the activation dimension), while \citet{cunningham2026} train multi-layer probes, leading to a storage cost and loading time that is additionally proportional to the number of layers in the model.

We use TPU machines to dump training, validation, and test activations to disk (training, val, and test), and then use CPUs for evaluating probes on these activations. We also are able to use small single-host TPU setups for training the probes (instead of much larger deployments that can host an entire production Gemini model).

Another infrastructure consideration is performing evaluations on long-context data. Our long-context evaluation dataset does not fit on one TPU node. We thus run long context evaluators distributed on many CPU workers, all of which load a random split of the data themselves. This is similar to typical single program multiple data setups \citep{darema1988single}. 


\textbf{Specific Considerations For Long Context Training:}
To train probes on long context data, it is impossible to load all training data at once to the TPU memory due to the prohibitively large data size of training activations. Therefore, for each epoch, we need to distribute both short context and long context data across data loading batches, and further divide into smaller training or evaluation batches. Training attention probes with long context data thus requires smaller batch sizes to more efficiently leverage data distributions and converge under smaller number of epochs. Moreover, we implemented a round-robin data allocation policy of data loading batches between training datasets, to prevent training dataset distribution shifts within an epoch. Loading large scale activations is expensive with standard training pipelines, so we implemented a customized pipeline with optimizations for data loading speed, padding, and pipelining.
These optimizations unlocked stable long context data probe training and improved accelerator utilization by about 5x.


\section{Error Bar Methodology for \Cref{figMainFigScatterCost}}
\label{appErrorBars}

The error bars in \Cref{figMainFigScatterCost} represent 95\% confidence intervals. We use fundamentally different uncertainty quantification methods for probes versus language models because they face different sources of variance:

\begin{itemize}
    \item \textbf{Probes}: We train 100 random seeds and select the best by validation loss. The dominant source of uncertainty is \textit{seed selection variance}: a different draw of 100 seeds would yield a different ``best'' model with different test performance.
    \item \textbf{Language Models}: We run a single evaluation with temperature zero, so there is no seed or sampling variance. The only uncertainty is \textit{finite test set noise}: our test set provides a noisy estimate of the true error rate.
\end{itemize}

These fundamentally different uncertainty sources require different statistical methods.

\subsection{Probe Methods (AlphaEvolve, Selected Probe, EMA Linear Probe)}

For probes, we use a smoothed bootstrap via kernel density estimation (KDE) in logit space to quantify seed selection uncertainty. Standard bootstrap of the argmin statistic produces degenerate confidence intervals because the same seed tends to be selected repeatedly. KDE smoothing addresses this by allowing the resampled seeds to have slightly different (val, test) pairs than the original data.

\textbf{Procedure:}
\begin{enumerate}
    \item Given $N$ trained seeds with validation losses $\{v_1, \ldots, v_N\}$ and corresponding test losses $\{t_1, \ldots, t_N\}$
    \item Transform test losses to logit space: $\ell_i = \log(t_i / (1 - t_i))$ to respect the bounded $[0,1]$ domain
    \item Fit a 2D kernel density estimate to the joint distribution of $(v_i, \ell_i)$ pairs
    \item Filter out degenerate seeds where $v_i$ equals the maximum penalty value (indicating failed training runs)
    \item For $B=20{,}000$ bootstrap iterations:
    \begin{enumerate}
        \item Sample $N$ synthetic $(v^*, \ell^*)$ pairs from the KDE
        \item Select $j^* = \argmin_k v^*_k$ (best validation seed in synthetic sample)  
        \item Transform back: $t^*_{j^*} = \sigma(\ell^*_{j^*})$ where $\sigma$ is the sigmoid function
        \item Record $t^*_{j^*}$ (test loss of selected seed)
    \end{enumerate}
    \item Report the 2.5th and 97.5th percentiles of the $B$ recorded test losses
\end{enumerate}

The KDE bandwidth is computed automatically using Scott's rule. This procedure captures how much the final reported test loss would vary if we repeated the entire training and selection process.

\textbf{Apparent One-Sided Error Bars:} Some probes exhibit error bars that appear one-sided, with a large downward extent but minimal upward extent. This occurs when the actual point estimate falls near the upper tail of the bootstrap distribution. Intuitively, this means the selected seed happened to perform relatively poorly on the test set compared to what would typically be expected. Equivalently, the selected seed landed near the 95th percentile of the bootstrap distribution rather than the median. In such cases, the confidence interval correctly indicates substantial room for improvement (downward) but limited room for the result to be worse (upward).

\subsection{Language Model Methods (Gemini 2.5 Flash)}

For the language model classifier, we use an analytic binomial confidence interval. Since we run only one deterministic evaluation (temperature zero), there is no seed variance to bootstrap over. Instead, we quantify uncertainty from the finite test set size.

Our error metric is a weighted combination of error rates across different test groups (see \Cref{eqnTradeoff}). For each group $g$ with weight $w_g$, we compute:
\begin{itemize}
    \item The empirical error rate $\hat{p}_g$ on $n_g$ test examples
    \item The binomial variance: $\text{Var}(\hat{p}_g) = \frac{\hat{p}_g(1-\hat{p}_g)}{n_g}$
\end{itemize}

The overall weighted error has variance:
\begin{equation}
    \text{Var}(\text{weighted error}) = \sum_g \left(\frac{w_g}{\sum_{g'} w_{g'}}\right)^2 \cdot \frac{\hat{p}_g(1-\hat{p}_g)}{n_g}
\end{equation}

We report the 95\% CI as $\text{mean} \pm 1.96 \cdot \sqrt{\text{Var}}$, clipped to be non-negative.

\subsection{Cascading Methods (Best Probe + Gemini 2.5 Flash)}

For the cascading classifier, we propagate uncertainties from both components. Let $f_{\text{defer}}$ be the deferral rate (fraction of examples sent to the LLM). The cascade error is approximately:
\begin{equation}
    \text{Cascade Error} \approx (1 - f_{\text{defer}}) \cdot \text{Probe Error} + f_{\text{defer}} \cdot \text{LLM Error}
\end{equation}

We propagate asymmetric uncertainties using the bootstrap CI widths from each component. Specifically, for the probe we use the distance from the bootstrap median to the 2.5th and 97.5th percentiles, and for the LLM we use the symmetric analytic margin. Combining via quadrature (assuming independence):
\begin{align}
    \delta_{\text{lo}} &= \sqrt{((1-f_{\text{defer}}) \cdot \delta^{\text{probe}}_{\text{lo}})^2 + (f_{\text{defer}} \cdot \delta^{\text{LLM}}_{\text{lo}})^2} \\
    \delta_{\text{hi}} &= \sqrt{((1-f_{\text{defer}}) \cdot \delta^{\text{probe}}_{\text{hi}})^2 + (f_{\text{defer}} \cdot \delta^{\text{LLM}}_{\text{hi}})^2}
\end{align}

where $\delta^{\text{probe}}_{\text{lo/hi}}$ are the lower and upper CI half-widths from the bootstrap distribution (not from the point estimate), and $\delta^{\text{LLM}}_{\text{lo/hi}}$ are the symmetric analytic margins. Since deferral rates are typically small ($\sim$8\%), the cascade uncertainty is dominated by the probe component, inheriting its asymmetric character.

\subsection{Interpretation}

The resulting confidence intervals have different interpretations:
\begin{itemize}
    \item \textbf{Probe CIs} answer: ``If we trained a different set of 100 seeds and picked the best, how much would our test error vary?''
    \item \textbf{LLM CIs} answer: ``Given our finite test set, how precisely do we know the true error rate?''
    \item \textbf{Cascade CIs} answer: ``What is the combined uncertainty from both probe seed selection and LLM test set noise?''
\end{itemize}

Probe CIs are typically wider than LLM CIs because seed selection introduces more variance than test set sampling given our test set sizes.

\section{Discusssion and results from weighting \Cref{eqnTradeoff} differently}
\label{appDifferent1250Weighting}

In earlier iteration in this paper, we used a weighting of 1 rather than 5 in \Cref{eqnTradeoff}. This appeared to lead to lower false positives, but extremely high false negative rates, and so we changed it once to the metric used in the main paper (we also sometimes measured a metric with hard cutoffs for false positive rates, but found this too confusing due to discontinuities). However, for completeness and transparency we include the results for the 1:2:50 weighting in \Cref{tabAppendix1250Weighting}.

\begin{table}[ht]
    \centering
    \caption{Results with 1:2:50 weighting in \Cref{eqnTradeoff}. Format matches \Cref{tabMainResults}.}
    \label{tabAppendix1250Weighting}

\footnotesize
\setlength{\tabcolsep}{3pt}
\arrayrulecolor{gray!30}
\begin{tabular}{p{3cm}cccc|ccccc|c}
\toprule
\arrayrulecolor{black}
& \multicolumn{4}{c|}{\textbf{FPR} $\downarrow$} & \multicolumn{5}{c|}{\textbf{FNR} $\downarrow$} & \\
\cmidrule(lr){2-5} \cmidrule(lr){6-10}
Classifier & SC[OT] & SC[HN] & LC[RT] & MT[HN] & SC[A] & LC[A] & MT[A] & SC[J] & SC[ART] & \textbf{Test Error} $\downarrow$ \\
\midrule
\arrayrulecolor{gray!30}
Rolling Attn Probe (20H), Rolling Agg & 0.11\% & 4.03\% & 0.37\% & \textbf{0.00\%} & 17.63\% & 14.72\% & 41.31\% & 6.20\% & 69.84\% & \textbf{0.87\%} \\ \hline
Rolling Attn Probe, Rolling Agg & 0.21\% & \textbf{2.01\%} & 0.37\% & 1.64\% & 16.71\% & 9.96\% & 39.38\% & 7.67\% & 69.61\% & 0.88\% \\ \hline
Vanilla Attn & 0.08\% & 3.36\% & \textbf{0.00\%} & \textbf{0.00\%} & 19.49\% & 92.21\% & 59.85\% & \textbf{1.65\%} & 47.56\% & 0.94\% \\ \hline
Gemini 2.5 Pro & 0.20\% & 6.71\% & 0.25\% & 13.11\% & 16.01\% & 13.42\% & 20.85\% & 10.64\% & 33.87\% & 0.94\% \\ \hline
Rolling Attn Probe, MultiMax Agg & \textbf{0.04\%} & 6.04\% & 0.49\% & 6.56\% & 20.65\% & 4.76\% & 35.14\% & 8.02\% & 53.13\% & 0.95\% \\ \hline
AlphaEvolve & 0.05\% & \textbf{2.01\%} & 0.25\% & \textbf{0.00\%} & 23.67\% & 30.30\% & 45.17\% & 47.28\% & 60.09\% & 0.96\% \\ \hline
AlphaEvolve (Early Variant) & 0.07\% & 4.70\% & 0.37\% & 1.64\% & 19.95\% & 8.23\% & 29.34\% & 51.51\% & 59.40\% & 0.96\% \\ \hline
Gemini 2.5 Flash & 0.25\% & 14.09\% & 0.74\% & 19.67\% & \textbf{8.58\%} & 6.06\% & 13.51\% & 8.74\% & \textbf{20.65\%} & 1.32\% \\ \hline
Linear Probe Mean & 0.06\% & 2.68\% & \textbf{0.00\%} & 3.28\% & 29.93\% & 100.00\% & 93.44\% & 89.06\% & 83.06\% & 1.63\% \\ \hline
Gemini 2.5 Flash Lite & 0.75\% & 25.50\% & 0.25\% & 14.75\% & 11.37\% & 55.41\% & 27.41\% & 10.64\% & 41.76\% & 1.78\% \\ \hline
MultiMax Trained, MultiMax Agg & 0.18\% & 2.68\% & 3.33\% & 3.28\% & 16.71\% & 3.90\% & 37.45\% & 22.03\% & 71.93\% & 2.34\% \\ \hline
MultiMax (Attn Trained) & 0.13\% & 7.38\% & 3.82\% & 24.59\% & 24.13\% & 2.16\% & 16.22\% & 9.28\% & 41.53\% & 2.82\% \\ \hline
Linear Probe Rolling Exp & 0.35\% & 12.08\% & 4.19\% & 67.21\% & 18.10\% & 4.76\% & \textbf{13.13\%} & 6.76\% & 26.91\% & 3.90\% \\ \hline
Rolling Attn Probe (20H), MultiMax Agg & 0.21\% & 5.37\% & 19.33\% & 24.59\% & 21.58\% & \textbf{0.00\%} & 23.94\% & 6.34\% & 25.06\% & 10.07\% \\ \hline
MultiMax Trained (20H), MultiMax Agg & 0.14\% & 4.70\% & 23.40\% & 1.64\% & 18.33\% & 0.87\% & 36.29\% & 57.39\% & 81.21\% & 11.96\% \\
\arrayrulecolor{black}
\bottomrule
\end{tabular}


\end{table}

\section{Attention Probe Inference}
\label{appAttentionProbeInference}


One might think that attention probes have quadratic computational costs in the sequence length, similar to standard transformers. However, this is not the case: we present an attention probe inference algorithm that scales as $O(n)$ in the sequence length $n$ (\Cref{secNotation}), at least for attention probes with a constant query vector $\vq$, as in \Cref{eqnAttentionProbe} (but note that this argument doesn't hold for attention probes with multiple layers of attention pooling).

The key insight is that we can derive a recursive update rule that allows attention probes to be updated incrementally during generation without storing a KV cache or recomputing attention scores over previous tokens.

Using the notation from \Cref{subsubsecAttention}, let the MLP-transformed activation at step $n$ be $\vy_n$. For a single head with query vector $\vq$ and value vector $\veevee$, the unnormalized attention score is $s_n = \vq^\top \vy_n$ and the value is $v_n = \veevee^\top \vy_n$. The probe output after $n$ tokens, denoted $A_n$, is the softmax-weighted average:

\begin{equation}
    A_n = \frac{\sum_{j=1}^{n} e^{s_j} v_j}{\sum_{j=1}^{n} e^{s_j}}
\end{equation}

\textbf{Recursive Update.} Let $Z_n = \sum_{j=1}^{n} e^{s_j}$ be the denominator. When a new token $n+1$ arrives, we update the denominator: $Z_{n+1} = Z_n + e^{s_{n+1}}$. We can express the new weighted average $A_{n+1}$ in terms of the previous state $A_n$:

\begin{align}
    A_{n+1} &= \frac{1}{Z_{n+1}} \left( \sum_{j=1}^{n+1} e^{s_j} v_j \right) \\
            &= \frac{1}{Z_{n+1}} \left( Z_n A_n + e^{s_{n+1}} v_{n+1} \right) \\
            &= A_n \left( \frac{Z_n}{Z_{n+1}} \right) + v_{n+1} \left( \frac{e^{s_{n+1}}}{Z_{n+1}} \right)
\end{align}

Let $\\beta_{n+1} = e^{s_{n+1}} / Z_{n+1}$ be the effective attention weight of the new token. Note that $Z_n / Z_{n+1} = 1 - \\beta_{n+1}$. The update simplifies to a standard exponential moving average form with a variable decay rate:

\begin{equation}
    A_{n+1} = A_n + \beta_{n+1} (v_{n+1} - A_n)
\end{equation}

\textbf{Numerical Stability.} To prevent overflow of $Z_n$, we track the log-sum-exp, $\ell_n = \log Z_n$. The update for the log-normalizer is:
\begin{equation}
    \ell_{n+1} = \log(e^{\ell_n} + e^{s_{n+1}}) = m + \log(e^{\ell_n - m} + e^{s_{n+1} - m})
\end{equation}
where $m = \max(\ell_n, s_{n+1})$. The weight $\beta_{n+1}$ is then computed stably in log-space:
\begin{equation}
    \beta_{n+1} = \exp(s_{n+1} - \ell_{n+1})
\end{equation}

\textbf{Conclusion.} To run an attention probe during generation, we only need to maintain the current scalar average $A_n$ and the scalar log-normalizer $\ell_n$ per head. Upon generating token $n+1$, we compute $s_{n+1}$ and $v_{n+1}$, update $\ell_{n+1}$ and $A_{n+1}$ using the equations above, and discard the token activations. This gives $O(n)$ total compute and $O(1)$ memory for a sequence of length $n$.

It is clear how this approach can be generalized to many heads, and architectures such as the Max of Rolling Means Attention Probe (\Cref{subsubsecRollingMeans}) can also have $O(n)$ efficiency by storing all the window scores (in our case the window is size 10, so small), and taking the max of the current rolling sum and the prior max.

\end{document}